\documentclass[a4paper]{article}

\usepackage[english]{babel}
\usepackage[utf8]{inputenc}
\usepackage[T1]{fontenc}

\usepackage[a4paper,top=3cm,bottom=2cm,left=3cm,right=3cm,marginparwidth=1.75cm]{geometry}

\usepackage{amsmath}
\usepackage{graphicx}
\usepackage[colorinlistoftodos]{todonotes}
\usepackage[colorlinks=true, allcolors=blue]{hyperref}

\usepackage{amssymb}
\usepackage{booktabs}
\usepackage{dsfont}
\usepackage{multirow}
\usepackage{parskip}

\DeclareMathOperator*{\argmax}{arg\,max}

\newcommand*\rot{\rotatebox{90}}

\title{End-to-end semantic face segmentation with conditional random fields as convolutional, recurrent and adversarial networks}
\author{%
Umut~Güçlü\textsuperscript{*, 1},
Yağmur~Güçlütürk\textsuperscript{*, 1},\\
Meysam~Madadi\textsuperscript{2},
Sergio~Escalera\textsuperscript{3},
Xavier~Baró\textsuperscript{4},
Jordi~González\textsuperscript{2},\\
Rob~van~Lier\textsuperscript{1},
Marcel~van~Gerven\textsuperscript{1}
}

\date{}

\begin{document}
\maketitle

\begin{abstract}
Recent years have seen a sharp increase in the number of related yet distinct advances in semantic segmentation. Here, we tackle this problem by leveraging the respective strengths of these advances. That is, we formulate a conditional random field over a four-connected graph as end-to-end trainable convolutional and recurrent networks, and estimate them via an adversarial process. Importantly, our model learns not only unary potentials but also pairwise potentials, while aggregating multi-scale contexts and controlling higher-order inconsistencies. We evaluate our model on two standard benchmark datasets for semantic face segmentation, achieving state-of-the-art results on both of them.
\end{abstract}

\renewcommand*{\thefootnote}{\fnsymbol{footnote}}
\footnotetext[1]{Umut Güçlü and Yağmur Güçlütürk contributed equally to this work.\\}
\renewcommand*{\thefootnote}{\arabic{footnote}}
\footnotetext[1]{Radboud University, Donders Institute for Brain, Cognition and Behaviour, Nijmegen, the Netherlands}
\footnotetext[2]{Universitat Autònoma de Barcelona and Computer Vision Center, Barcelona, Spain}
\footnotetext[3]{Universitat de Barcelona and Computer Vision Center, Barcelona, Spain}
\footnotetext[4]{Universitat Oberta de Catalunya and Computer Vision Center, Barcelona, Spain}

\section{Introduction}
\label{section:introduction}

Semantic segmentation is a very important topic in computer vision primarily because of its many applications in object recognition, image annotation, image coding, scene understanding and biomedical image processing. One specific field of semantic segmentation is face segmentation in which the task is to correctly assign labels of face regions such as nose, mouth, eye, hair, etc. to each pixel in a face image. Face segmentation techniques are frequently used in security systems and in the field of human computer interaction, mainly in order to facilitate the problems of face detection and recognition~\cite{pujol2017face, ahonen2006face, mian2007an, luu2016a}, and emotion/expression recognition~\cite{cohen2000emotion, stathopoulou2010on, happy2015automatic}. Further specialized entertainment oriented applications of face segmentation include style transfer~\cite{elad2016style-transfer}, virtual make-up application~\cite{liu2016makeup}, virtual face-swapping~\cite{korshunova2016fast} and 3D performance capturing~\cite{saito2016real-time}. Additionally, utilization of face segmentation techniques could potentially improve the performance in several other computer vision tasks involving the processing of face images such as apparent personality prediction~\cite{gucluturk2016deep} and face hallucination~\cite{gucluturk2016convolutional}.

Semantic segmentation of faces is a difficult problem because of the large number of variable conditions that need to be considered, especially when applied to face pictures taken in uncontrolled environments. These conditions include variations in facial expression, skin color, lighting, image quality, pose, hair texture and style, as well as the presence of varying amounts of background clutter and occlusions. Furthermore, despite extensive studies in face segmentation, correctly classifying hair pixels still remains a particularly challenging task~\cite{wang2011a}, largely due to the inherent properties of hair such as color similarity to background, non-rigidity and non-unique shape.

Recently, there has been a sizable number of advances in semantic segmentation. In the context of semantic image segmentation,~\cite{zheng2015conditional} showed that formulating the iterative update equation of a CRF over a fully-connected graph~\cite{krahenbuhl2012efficient} as a recurrent neural network (RNN) resulted in state-of-the-art accuracy on Pascal VOC 2012 dataset~\cite{everingham2014the}. While this model did not learn the pairwise potential of the CRF and relied on fixed Gaussian kernels, it was end-to-end trainable. In the context of semantic face segmentation,~\cite{liu2015multi-objective} showed that formulating the unary potential and the pairwise potential of a conditional random field (CRF) over a four-connected graph as a convolutional neural network (CNN) resulted in state-of-the-art accuracy on the Part Labels dataset~\cite{learned-Miller2016labeled, kae2013augmenting} and the Helen dataset~\cite{le2012interactive, smith2013exemplar-based}. While this model was not end-to-end trainable and relied on graph cuts, it learned both the the unary potential and the pairwise potential of the CRF.

Furthermore,~\cite{yu2015multi-scale} showed that the results of convolutional semantic segmentation models can be improved by using dilated kernels instead of regular kernels, which increase receptive field size without decreasing receptive field resolution. Similarly,~\cite{luc2016semantic} showed that results of convolutional semantic segmentation models can be improved by using an adversarial loss function in addition to a segmentation loss function, which enforces higher-order consistencies without explicitly taking into account any higher-order potentials.

Here, our goal is to formulate a model for semantic face segmentation by combining the respective strengths of the aforementioned models. That is, the model should be end-to-end trainable like~\cite{zheng2015conditional}, and learn both the unary potential and the pairwise potential of a CRF over a four-connected graph like~\cite{liu2015multi-objective} while aggregating multi-scale contexts like~\cite{yu2015multi-scale} and controlling higher-order inconsistencies like~\cite{luc2016semantic}. Table~\ref{table_1} shows an overview of the differences and the similarities between our model (i.e., CnnRnnGan), its variants (i.e., Cnn, CnnGan and CnnRnn), and the recent models that they are based on.

\begin{table}[]
\centering
\caption{\textbf{An overview of the differences and the similarities between the variants of our model, and the recent semantic segmentation models that they are based on.} $\psi_u$ and $\psi_p$ denote learned instead of fixed unary potentials and pairwise potentials, respectively.}
\label{table_1}
\begin{tabular}{@{}lccccc@{}}
\toprule
 & \multirow{2}{*}{\begin{tabular}[c]{@{}c@{}}adversarial\\ training\end{tabular}} & \multicolumn{3}{c}{conditional random field} & \multirow{2}{*}{\begin{tabular}[c]{@{}c@{}}dilated\\ conv.\end{tabular}} \\
 &  & ($\psi_u$) & ($\psi_p$) & (end-to-end) &  \\ \midrule
Yu and Koltun (2015)~
 \cite{yu2015multi-scale}& $\times$ & ----- & ----- & ----- & \checkmark \\
Liu et al. (2015)~
\cite{liu2015multi-objective}& $\times$ & \checkmark & \checkmark & $\times$ & $\times$ \\
Zheng et al. (2015)~
\cite{zheng2015conditional}& $\times$ & \checkmark & $\times$ & \checkmark & $\times$ \\
Luc et al. (2016)~
\cite{luc2016semantic}& \checkmark & ----- & ----- & ----- & \checkmark \\ \midrule
Cnn (Ours) & $\times$ & ----- & ----- & ----- & \checkmark \\
CnnGan (Ours) & \checkmark & ----- & ----- & ----- & \checkmark \\
CnnRnn (Ours) & $\times$ & \checkmark & \checkmark & \checkmark & \checkmark \\
CnnRnnGan (Ours) & \checkmark & \checkmark & \checkmark & \checkmark & \checkmark \\ \bottomrule
\end{tabular}
\end{table}

The contributions of our work are the following:
\begin{enumerate}
\item We propose an end-to-end trainable convolutional and recurrent network formulation of a conditional random field over a four-connected graph with learnable unary potentials and pairwise potentials in which dilated convolutions and adversarial training are used for aggregating multi-scale contexts and controlling higher-order inconsistencies, respectively.
\item We exploit the structured nature of faces by conditioning the model on face landmarks, and/or training multiple models for different face landmarks and combining their outputs akin to part-based models.
\item We evaluate the model on two standard semantic face segmentation datasets (i.e., Part Labels and Helen), achieving state-of-the-art results on both of them while considerably improving the segmentation accuracy of challenging face parts such as hair.
\end{enumerate}

The rest of this paper is organized as follows: In the next section, we overview the recent work on semantic segmentation in general and semantic face segmentation in particular. In Section~\ref{section:methods}, we present our model. In Section~\ref{section:results}, we present the results of the main experiments, in which we evaluate our model on both the Part Labels dataset and the Helen dataset, compare the obtained results versus the state-of-the-art, and present the results of the ablation experiments, in which we evaluate variants of our model on the same datasets. In the last section, we conclude with an overview of our work.

\section{Related work}
\label{section:related_work}

Semantic segmentation has been widely studied in computer vision in a wide spectrum of domains. For a comprehensive review of classical approaches for semantic segmentation, we refer the reader to~\cite{zhu2016beyond}. In this section, we review recent work on semantic segmentation in general and semantic face segmentation in particular. 

The most recent state-of-the-art semantic segmentation models almost exclusively rely on convolutional neural networks. In contrast to earlier approaches where recognition architectures were directly used for semantic segmentation~\cite{farabet2013learning}, current approaches utilize architectures that are carefully adapted for the task at hand. \cite{long2015fully} proposed the first such approach, where the fully-connected layers of popular architectures such as AlexNet~\cite{krizhevsky2012imagenet}, VGGNet~\cite{simonyan2014very} and GoogLeNet~\cite{szegedy2014going} were replaced with (de)convolution layers and combined with earlier layers to enable dense and high resolution predictions. Since then, this approach has been continuously improved by the introduction of more sophisticated architectures, which enabled denser~\cite{Noh_2015_ICCV, badrinarayanan2015segnet, hong2015decoupled}, higher resolution predictions~\cite{ ghiasi2016laplacian}, and/or encoder-decoder~\cite{zheng2015learning} predictions. In particular, \cite{yu2015multi-scale} proposed dilated convolutions for dense prediction, where contexts could be aggregated by multiscale levels without loss of neither resolution nor coverage. This idea has been extended by~\cite{chen2016deeplab} to enable a larger field of view through spatial pyramid pooling. Such approaches enjoy the benefits of dense and high resolution predictions without the burden of extra parameters.

At the same time, conditional random fields have been used in semantic segmentation for postprocessing outputs of region-level or pixel-level semantic segmentation models. While the relatively small number of outputs of region-based semantic segmentation models could be postprocessed by CRFs with dense pairwise connectivity~\cite{kae2013augmenting}, the relatively large number of outputs of pixel-level semantic segmentation models could only be postprocessed by CRFs with sparse pairwise connectivity~\cite{liu2015multi-objective}. In a seminal work,~\cite{krahenbuhl2012efficient} proposed an efficient iterative algorithm for approximate inference in fully-connected CRFs with Gaussian edge potentials, which has been widely adopted for postprocessing outputs of pixel-level segmentation models~\cite{Noh_2015_ICCV, badrinarayanan2015segnet, hong2015decoupled}. \cite{zheng2015conditional} formulated this algorithm as a recurrent neural network, which is trained along with a pixel-level segmentation model instead of postprocessing it. This formulation is reminiscent of the pixel-level semantic segmentation model in~\cite{pinheiro2014recurrent}, whose outputs were iteratively refined with a recurrent convolutional neural network.

Recently, generative adversarial networks (GANs)~\cite{goodfellow2014generative} have received particular attention in computer vision~\cite{radford2015unsupervised, denton2015deep, chen2016infogan}. The idea behind GANs is training a discriminator and a generator by letting them play a two-player minimax game. In this game, the objective of the discriminator is distinguishing samples that are drawn from the data distribution from samples that are drawn from the model distribution, and the objective of the generator is fooling the discriminator. While GANs have been proposed for estimating generative models via an adversarial process, they have been widely adopted for other tasks such as inpainting~\cite{pathak2016context}, style transfer~\cite{li2016precomputed} and super-resolution~\cite{ledig2016photo-realistic} as loss functions. In particular,~\cite{luc2016semantic} estimated a semantic segmentation model via an adversarial process by training a discriminator for distinguishing ground-truths from outputs of the semantic segmentation model and the semantic segmentation model for fooling the discriminator. They showed that this process leads to improved results on the Stanford Background dataset~\cite{gould2009decomposing} and the PASCAL VOC 2012 dataset~\cite{everingham2014the}.

There has been relatively fewer semantic face segmentation models that rely on convolutional neural networks. Most earlier models were based on CRFs~\cite{kae2013augmenting}, hand designed features~\cite{warrell2009labelfaces} and exemplars~\cite{smith2013exemplar-based}. Kae et al.~\cite{kae2013augmenting} modeled global part dependencies using a restricted Boltzmann machine to have an overall realistic shape while local shape details were modeled through a CRF, whereas Smith et al.~\cite{smith2013exemplar-based} used exemplar-based non-rigid warping for face segmentation. Despite the progress in the models, hair segmentation is still the most challenging part due to its color and style variability. Earlier works include attempts of modeling hair, skin and background color \cite{scheffler2011joint, yacoob2006detection}, mixture of hair styles~\cite{lee2008markov} or MRF/CRF labeling~\cite{huang2008towards}. As a specialized hair segmentation, Wang et al.~\cite{wang2012good} applied co-occurrence probabilities of face components identified by a Markov random field. The final segmentations were constrained in a tree-structured model built over part co-occurrences. Among CNN-based face segmentation methods,~\cite{luo2012hierarchical} segmented faces based on a hierarchical part detection process, where the face was detected as the root of the hierarchy and the smallest components of the face were detected at the bottom of such hierarchy. Then, an autoencoder network transformed those detected components into label maps. As a result of using hierarchies, partially occluded faces could be easily handled. Recently,~\cite{zhou2015interlinked} applied the part detection idea by training one network for each part and mapping the segmentation result to the original image. \cite{liu2015multi-objective} generated pairwise terms as class edge potentials through a four-connected graph. Such edge potentials extracted in a multi-objective network along with unary terms were trained using non-structured loss functions, and provided prior knowledge to the network by including inaccurate segmentations as an additional network input. This study showed the benefits of including prior knowledge for improving face segmentation.

\section{End-to-end semantic face segmentation}
\label{section:methods}

\begin{figure*}
\centering
\includegraphics[width=1\textwidth]{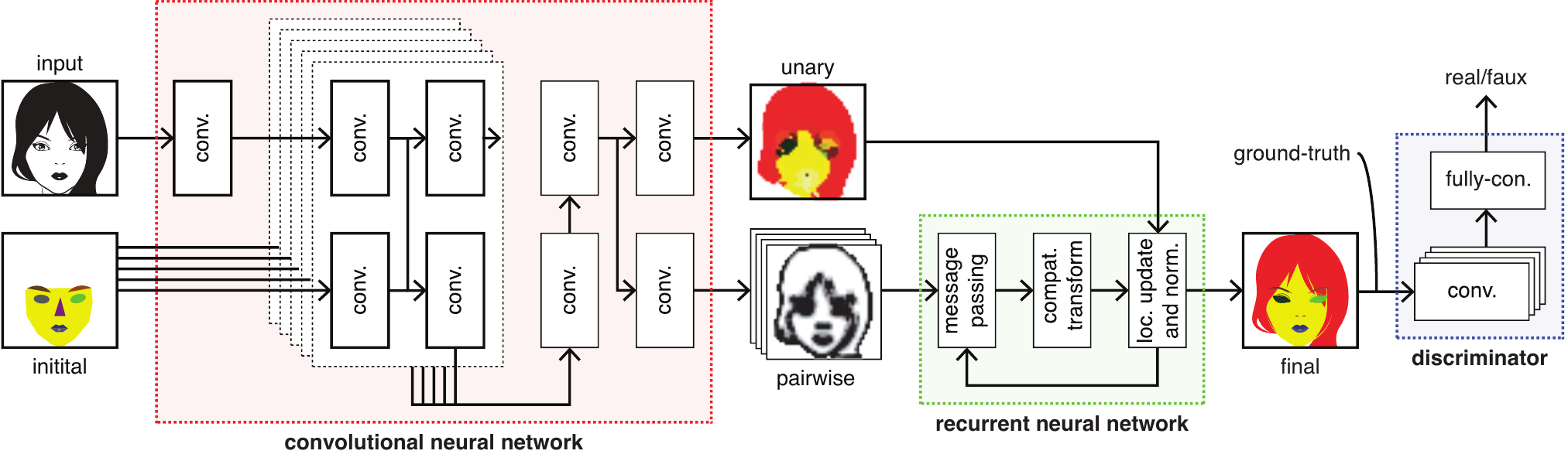}
\caption{
\label{figure:figure_1}
\textbf{Our semantic face segmentation model.} The conditional random field is formulated as a composition of two neural networks:
i) A convolutional neural network, which nonlinearly transforms an input face and its initial segmentation to the unary potential and the pairwise kernels of the conditional random field.
ii) A recurrent neural network, which transforms the unary potential and the pairwise kernels of the conditional random field to the final segmentation of the input face.
In the training phase, a discriminator and the conditional random field play a two-player minimax game, in which the objective of the discriminator is distinguishing ground-truth segmentations from final segmentations, and the objective of the conditional random field is fooling the discriminator.
}
\end{figure*}

For end-to-end semantic face segmentation, we formulate a conditional random field as a composition of a convolutional neural network and a recurrent neural network (Section~\ref{section:conditional_random_field}). The convolutional neural network is used for obtaining the unary potential and the pairwise kernels of the conditional random field as a function of an input face and its initial segmentation (Section~\ref{section:convolutional_neural_network}). The recurrent neural network is used for obtaining the label compatibility function and a mean field approximation of the Gibbs distribution of the conditional random field as a function of the unary potential and the pairwise kernels of the conditional random field (Section~\ref{section:recurrent_neural_network}). In the training phase, a discriminator and the conditional random field play a two-player minimax game, in which the objective of the discriminator is distinguishing ground-truth segmentations from final segmentations, and the objective of the conditional random field is fooling the discriminator (Section~\ref{section:adversarial_training}). Fig.~\ref{figure:figure_1} illustrates our model. The following sections present the components of our model in detail.

\begin{figure}
\centering
\includegraphics[width=1\textwidth]{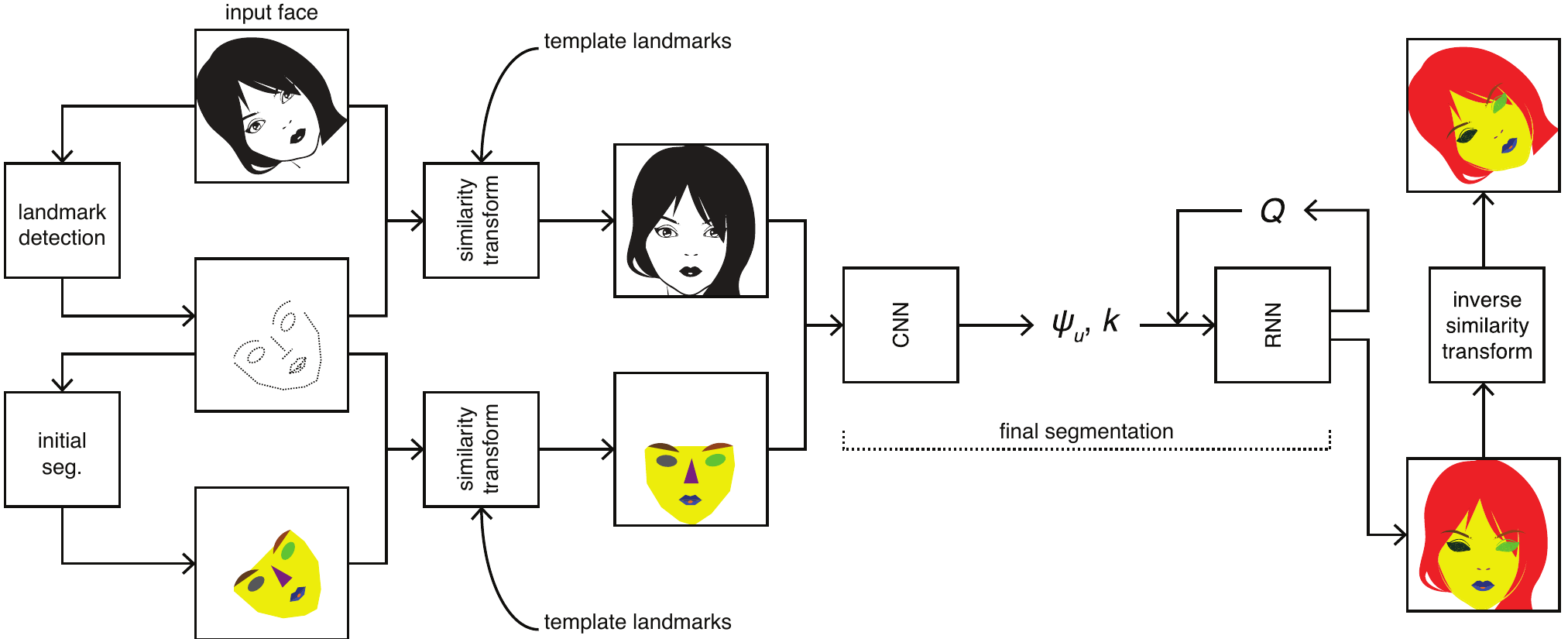}
\caption{\label{figure:figure_2}\textbf{Our semantic face segmentation pipeline.} 1. Sixty-eight landmarks of the the input face are detected. 2. An initial segmentation of the input face is obtained. 3. The input face and the initial segmentation of the input face are warped to the template face by using a similarity transformation, and resized to 500 pixels $\times$ 500 pixels. 4. The final segmentation of the face is obtained by using our model. 5. Optionally, the final segmentation of the input face can be resized back from 500 pixels $\times$ 500 pixels and warped back from the template face by using the inverse of the similarity transformation.}
\end{figure}

Prior to entering the model, an input face is preprocessed as follows: A template face is obtained by averaging the faces in the training set of the Part Labels dataset. Sixty-eight landmarks of the template face and the input face are detected by using the dlib implementation~\cite{king2009dlib-ml} of an ensemble of regression trees~\cite{kazemi2014one}.\footnote{Note that the dataset that was used for training the landmark detection model provided by dlib contains some of the images that we use to test our final segmentation model. To avoid circular analysis, we retrained the landmark detection model on the same dataset that it was originally trained on after removing these images.} An initial segmentation of the input face is obtained by filling the regions that are formed by connecting the landmarks around background, face skin, left eyebrow, right eyebrow, left eye, right eye, nose, upper lip, inner mouth and lower lip. A similarity transformation from the landmarks of the input face to the landmarks of the template face is estimated. The input face and its initial segmentation are warped to the template face by using the similarity transformation, and resized to 500 pixels $\times$ 500 pixels. The final segmentation of the input face is obtained by using our model. Optionally, the final segmentation of the input face can be resized back from 500 pixels $\times$ 500 pixels and warped back from the template face by using the inverse of the similarity transformation. Fig.~\ref{figure:figure_2} illustrates our preprocessing pipeline.

\subsection{Conditional random field}
\label{section:conditional_random_field}

We begin the exposition of our model by considering a conditional random field over a four-connected graph. Let $\mathbf{I} = \{I_1, \ldots, I_N\}$ and $\mathbf{X} = \{X_1, \ldots, X_N\}$ be random fields, where $I_i \in \mathds{R} ^ 3$ and $X_i \in \mathcal{L} = \{l_1, \ldots, l_P\}$ are the color vector and the label of the pixel $i \in \{1, \ldots, N = h \times w\}$, respectively. Let $\mathcal{G} = (\mathcal{V}, \mathcal{E})$ be a four-connected graph, where $\mathcal{V}$ contains all pixels, and $\mathcal{E}$ contains all pixel pairs that have a taxicab metric of one.

The conditional random field $(\mathbf{I}, \mathbf{X})$ over $\mathcal{G}$ is defined by the following Gibbs distribution:

\begin{equation}
P\left(\mathbf{X} = \mathbf{x} | \mathbf{I}\right) = \frac{1}{Z\left(\mathbf{I}\right)}\exp\left(-E\left(\mathbf{x} | \mathbf{I}\right)\right)
\end{equation}

where $Z$ is the partition function, and $E$ is the following Gibbs energy:

\begin{equation}
E\left(\mathbf{x}\right) = \sum_{i \in \mathcal{V}}\psi_u\left(x_i\right) + \sum_{i, j \in \mathcal{E}}\psi_p\left(x_i, x_j\right)
\end{equation}

where $\psi_u$ is the unary potential, which is the cost of assigning the label $x_i$ to the pixel $i$, and $\psi_p$ is the pairwise potential, which is the cost of assigning the labels $x_i$ and $x_j$ to the pixels $i$ and $j$, respectively. Note that we omit conditioning on $\mathbf{I}$ for notational convenience. The pairwise potential is of the following form:

\begin{equation}
\psi_p\left(x_i, x_j\right) = \mu\left(x_i, x_j\right)k_{i, j}
\end{equation}

where $\mu$ is a label compatibility function, which is not assumed to be symmetric since it was shown that this assumption improves semantic segmentation results~\cite{zheng2015conditional}, and $k$ is arbitrary pairwise kernels.

Following~\cite{krahenbuhl2012efficient}, we approximate the Gibbs distribution with the mean field distribution that minimizes the Kullback–Leibler divergence between the Gibbs distribution and the distributions that are of the following form:\footnote{While the Gibbs energy can be converted to a submodular energy, which makes exact inference (e.g. with combinatorial min cut/max flow algorithms) possible, we resort to approximate inference (i.e. with mean field theory) to be able to formulate it as a recurrent neural network, which makes end-to-end training possible.}

\begin{equation}
Q\left(\mathbf{X}\right) = \prod_{i \in \mathcal{V}}Q_i\left(X_i\right)
\end{equation}

This approximation results in the following iterative update equation:

\begin{equation}
Q_i\left(x_i = l\right) = \frac{1}{Z_i} \exp \left(-\psi_u\left(x_i\right) - \sum_{l' \in \mathcal{L}} \mu\left(l, l'\right) \sum_{i, j \in \mathcal{E}} k_{i, j} Q_j\left(l'\right) \right)
\end{equation}

\subsection{Convolutional neural network}
\label{section:convolutional_neural_network}

Following~\cite{liu2015multi-objective}, we formulate $\psi_u$ and $k$ as a convolutional neural network, whose architecture is inspired by recent architectures proposed in~\cite{yu2015multi-scale, vandenOord2016wavenet, kalchbrenner2016neural}.

The network comprises the following layers:
\begin{enumerate}
\item One convolution layer that has 32 kernels of size $3 \times 3$ with no nonlinearities.
\item Five blocks, where each block comprises the following layers:
\begin{enumerate}
\item Two parallel convolution layers that have 64 kernels of size $1 \times 1$ with no nonlinearities (i.e. bias layer) and 64 dilated kernels of size $3 \times 3$ with gated activation units~\cite{vandenOord2016wavenet} (i.e. weight layer). The input of the bias layer is the initial segmentation. The output of the bias layer is summed with the activation of the weight layer. The output of the weight layer becomes the input of the next layer.
\item Two parallel convolution layers that have 64 kernels of size $1 \times 1$ with no nonlinearities (i.e. residual layer) and 64 kernels of size $1 \times 1$ with rectified linear units (i.e. skip layer). The output of the residual layer is summed with the input of the block, which becomes the input of the next layer. The output of the skip layer is concatenated with the outputs of the skip layers of the remaining blocks along the channel axis, which becomes the input of the next layer after the last block.
\end{enumerate}
\item One convolution layer that has 160 kernels of size $1 \times 1$ with rectified linear units.
\item Two parallel convolution layers that have $P$ kernels of size $1 \times 1$ with no nonlinearities (i.e., $\psi_u$) and four kernels of size $1 \times 1$ with exponential units (i.e., $k$).
\end{enumerate}

Dilated kernels are the same as the regular kernels with the exception that successive kernel elements have holes between each other, whose size is determined by a dilation factor. As a result, they increase receptive field size without decreasing receptive field resolution. Note that regular convolution layers can be considered dilated convolution layers with a dilation factor of one.

The dilation factor of the first block is one, which is doubled after every block. The number of blocks (i.e., five) is chosen to be the largest possible value such that the receptive field dimensions of the last block is less than or equal to the pixel dimensions. That is:
\begin{equation}
q = \argmax_x f\left(x\right) : f\left(x\right) = 3 + 2\sum_{i = 0}^{x - 1}2^i \leq \min\left(h, w\right)
\end{equation}

where $q$ is the number of blocks.

\subsection{Recurrent neural network}
\label{section:recurrent_neural_network}

Following [3], we formulate $\mu$ and the iterative update equation as a recurrent neural network. The network comprises (i) a message passing layer, (ii) a compatibility transform layer, and (iii) a local update and normalization layer. Note that only the compatibility transform layer has free parameters. 

The layers are implemented as follows: Let $\psi_u$ be a $P \times h \times w$ tensor and $k$ be a $4 \times h \times w$ tensor, which are the outputs of the convolutional neural network. Prior to the first iteration, $Q$ is initialized with $\psi_u$, and the channels of $k$ are broadcasted to the shape of $Q$, which results in a set of four $P \times h \times w$ tensors.

\begin{itemize}
\item In the message passing layer, $Q$ is shifted up, right, down and left by one pixel, and multiplied (i.e. Hadamard product) with the corresponding elements of k, which results in a set of four $P \times h \times w$ tensors. The elements of this set are summed. As a result, this layer outputs a $P \times h \times w$ tensor, which becomes the input of the next layer.
\item In the compatibility transform layer, the input tensor is convolved with $P$ kernels of size 1 $\times$ 1. As a result, this layer outputs a $P \times h \times w$ tensor, which becomes the input of the next layer.
\item In the local update and normalization layer, the input tensor is subtracted from $-\psi_u$, exponentiated and normalized (i.e., softmax function). As a result, this layer outputs a $P \times h \times w$ tensor, which becomes the input of the first layer after the first four iterations and the output of the network after the fifth iteration.
\end{itemize}

\subsection{Adversarial training}
\label{section:adversarial_training}

While this end-to-end trainable convolutional and recurrent neural network formulation of a conditional random field can learn both the unary potential and the pairwise potential, it does not take into account any higher-order potentials that can enforce higher-order consistencies. To be able to enforce higher-order consistencies without explicitly taking into account any higher-order potentials, we train the model by minimizing an adversarial loss function in addition to a segmentation loss function~\cite{luc2016semantic}.

To this end, we train a discriminator along with our model, which is from now on referred to as the generator. We denote the output of the discriminator as $D_{\theta_D}(\mathbf{.})$, which is the probability that the input of the discriminator is a ground-truth segmentation. We denote the output of the generator as $G_{\theta_G}(\mathbf{.})$, which is the probabilities of assigning each of the $P$ labels to each of the $N$ pixels of the input.

In this context, the goal of the discriminator is to distinguish ground-truth segmentations from generated segmentations, whereas the goal of the generator is to generate segmentations that are indistinguishable from ground-truth segmentations. That is, they play the following minimax game:

\begin{equation}
\begin{aligned}
\min_{\theta_G}\max_{\theta_D} \quad & \mathds{E}_{\mathcal{T}^{\left(n\right)}\sim p_{\mathcal{T}}\left(\mathcal{T}^{\left(n\right)}\right)} \log D_{\theta_D}\left(\mathcal{T}^{\left(n\right)}\right) + \\
& \mathds{E}_{\mathcal{I}^{\left(n\right)}\sim p_{\mathcal{I}}\left(\mathcal{I}^{\left(n\right)}\right)} \log\left(1 - D_{\theta_D}\left(G_{\theta_G}\left(\mathcal{I}^{\left(n\right)}\right)\right)\right)
\end{aligned}
\end{equation}

where $\mathcal{I} = \{\mathcal{I}^{(1)}, \mathcal{I}^{(2)}, \ldots\}$ is a set of images, and $\mathcal{T} = \{\mathcal{T}^{(1)}, \mathcal{T}^{(2)}, \ldots\}$ is a set of corresponding ground-truth segmentations.

We formulate the discriminator as a convolutional neural network whose architecture is inspired by the architecture in~\cite{radford2015unsupervised}. The network comprises four convolution layers and a fully-connected layer. The $i$th convolution layer has $2 ^ {6 + i}$ kernels with a size of $3 \times 3$, a stride of $2 \times 2$, a pad of $1 \times 1$ and leaky rectified units~\cite{maas2013rectifier}. The activations of the first four convolution layers are normalized along the mini-batch (i.e., batch normalization~\cite{ioffe2015batch}). The output of the last convolution layer is averaged along the spatial axes (i.e., global average pooling~\cite{lin2013network}). The fully-connected layer has one kernel with a sigmoid unit.

The discriminator is trained by iteratively minimizing the following discriminator loss function:

\begin{equation}
L_{dis} = -\log D_{\theta_D}\left(\mathcal{T}^{\left(n\right)}\right) - \log\left(1 - D_{\theta_D}\left(G_{\theta_G}\left(\mathcal{I}^{\left(n\right)}\right)\right)\right)
\end{equation}

Note that $L_{dis}$ is the sum of two sigmoid cross entropy loss functions.

The generator is trained by iteratively minimizing the following linear combination of an adversarial loss function and a segmentation loss function:

\begin{eqnarray}
L_{gen} = L_{adv} + \lambda L_{seg}
\end{eqnarray}

where $\lambda$ is the coefficient of the segmentation loss function and the constituent loss functions are of the following forms:

\begin{eqnarray}
L_{adv} & = & -\log D_{\theta_D}\left(G_{\theta_G}\left(\mathcal{I}^{\left(n\right)}\right)\right) \\
L_{seg} & = & -\sum_{l \in \mathcal{L}}\sum_{i \in \mathcal{V}}\mathcal{T}^{\left(n\right)}_{l, i}\log G_{\theta_G}\left(\mathcal{I}^{\left(n\right)}\right)_{l, i}
\end{eqnarray}

Note that $L_{adv}$ is a sigmoid cross entropy loss function, and $L_{seg}$ is a softmax cross entropy loss function.

\section{Results}
\label{section:results}

\subsection{Implementation details}

The models were implemented in Chainer with CUDA and cuDNN~\cite{tokui2015chainer}.

The biases of the models were initialized with zero, the weights of the models were initialized with samples drawn from a scaled Gaussian distribution~\cite{he2015deep}, and the coefficient of the segmentation loss function (i.e., $\lambda$) was set to 100.

Adam~\cite{kingma2014adam} with initial $\alpha$ = 0.001, $\beta_1$ = 0.9, $\beta_2$ = 0.999 and $\epsilon$ = 1e-8 was used to iteratively train the models on the combination of the training set and the validation set\footnote{The hyperparameters (i.e., $\lambda$, $\alpha$ and the number of epochs) were optimized prior to combining the training set and the validation set.} for 111 epochs. The learning rate (i.e., $\alpha$) was reduced by a factor of 10 after 100 and 110 epochs.

At each iteration, the discriminator and the generator were updated sequentially. To prevent them from overpowering each other, the training of the discriminator was suspended or resumed if the following conditions were satisfied, respectively:

\begin{equation}
\frac{L_{dis}}{L_{adv}} < 0.1, \frac{L_{dis}}{L_{adv}} > 0.5
\end{equation}

Similarly, the training of the generator was suspended or resumed if the following conditions were satisfied, respectively:
\begin{equation}
\frac{L_{dis}}{L_{adv}} > 10, \frac{L_{dis}}{L_{adv}} < 2
\end{equation}

These conditions were selected based on~\cite{dosovitskiy2016generating}.

Our source code and pretrained models will be shared post-publication. Further details can be found at \url{https://github.com/umuguc}.

\subsection{Datasets}

We analyzed the Part Labels dataset and the Helen dataset in our experiments. These datasets are the standard benchmark datasets for semantic face segmentation, which comprise pairs of in-the-wild faces and ground-truth segmentations. Parts Label dataset comprises 2927 pairs of in-the-wild faces and ground-truth segmentations of background, face skin (including ear skin and neck skin) and hair (including facial hair), which is split in a 1500 pair training set, a 500 pair validation set and a 927 pair test set. Helen dataset comprises 2330 pairs of in-the-wild faces and ground-truth segmentations of face skin (excluding ear skin and neck skin), left eyebrow, right eyebrow, left eye, right eye, nose, upper lip, inner mouth, lower lip and hair (excluding facial hair), which is split in a 2000 pair training set, a 230 pair validation set and a 100 pair test set.

\subsection{Evaluation metrics}

Results are reported in terms of confusion matrix and Jaccard index (i.e., intersection over union). Confusion matrix is defined as the square matrix $\mathbf{A}$ where $A_{i, j}$ is the number of pixels whose true class is $i$ and predicted class is $j$. Jaccard index of class $i$ is defined as follows:

\begin{equation}
J_i = \frac{A_{i, i}}{\sum_jA_{i, j} + \sum_jA_{j, i} - A_{i, i}}
\end{equation}

Jaccard index of all classes is defined as follows:

\begin{equation}
J = \frac{\sum_iJ_i}{P}
\end{equation}

\subsection{Main experiments}

We conducted two main experiments on the Labeled Parts and the Helen datasets, in which we evaluated the CnnRnnGan model.

\subsubsection{Part Labels dataset}

We iteratively trained one global CnnRnnGan model for segmenting background, face skin and hair. Before the first iteration, the images in the dataset were resized to 106 pixels $\times$ 106 pixels. At each iteration, a mini-batch of size 16 was randomly selected without replacement, horizontally and vertically translated by $\pm$ 5 pixels, and mirrored in the left-right direction. Then, the mini-batch was cropped to the central 96 pixels $\times$ 96 pixels.

In the test phase, the inputs were oversampled (i.e., center and corners) and mirrored (i.e., left-right direction). The outputs were placed to their corresponding locations in the original inputs and averaged. Table~\ref{table_2} shows the resulting confusion matrix and Jaccard index. The most common cause of errors was mislabeling the classes as background. The least common cause of errors was mislabeling the classes as hair. All of the classes were segmented with a relatively high accuracy ($J = 0.8882$). Background was the most accurately segmented class ($J_{b} = 0.9656$). Hair was the least accurately segmented class ($J_{h} = 0.7808$).

\begin{table}[]
\centering
\caption{\textbf{The results of the main experiment on the Part Labels dataset.} Confidence matrix is reported in terms of percentage. The rest of the results are reported in terms of Jaccard index (i.e. intersection over union) of the classes and their arithmetic mean, respectively.}
\label{table_2}
\begin{tabular}{@{}cccccc@{}}
\toprule
 &  & \multicolumn{3}{c}{predicted class} &  \\
 &  & \includegraphics[width = 0.5cm, height = 0.5cm]{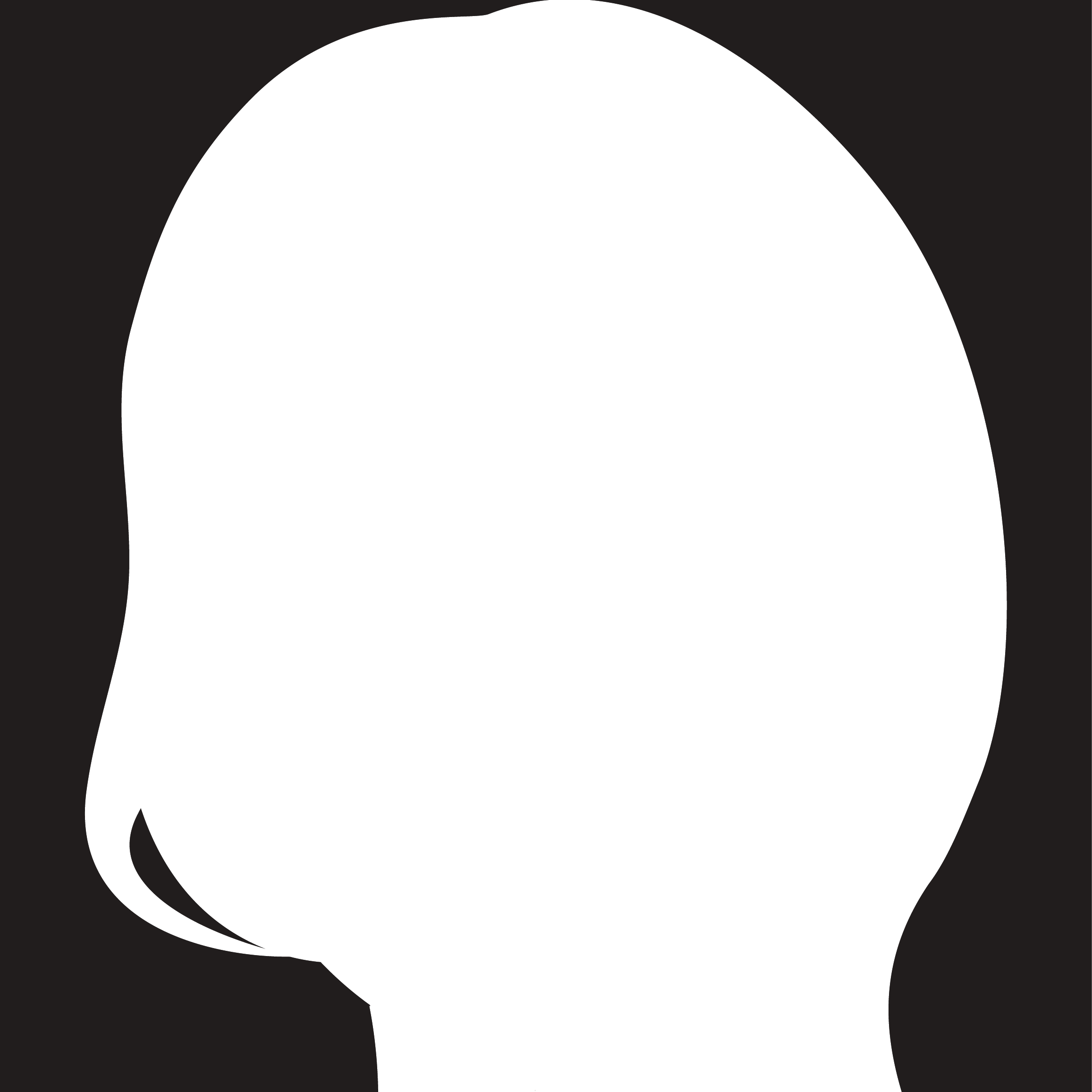} & \includegraphics[width = 0.5cm, height = 0.5cm]{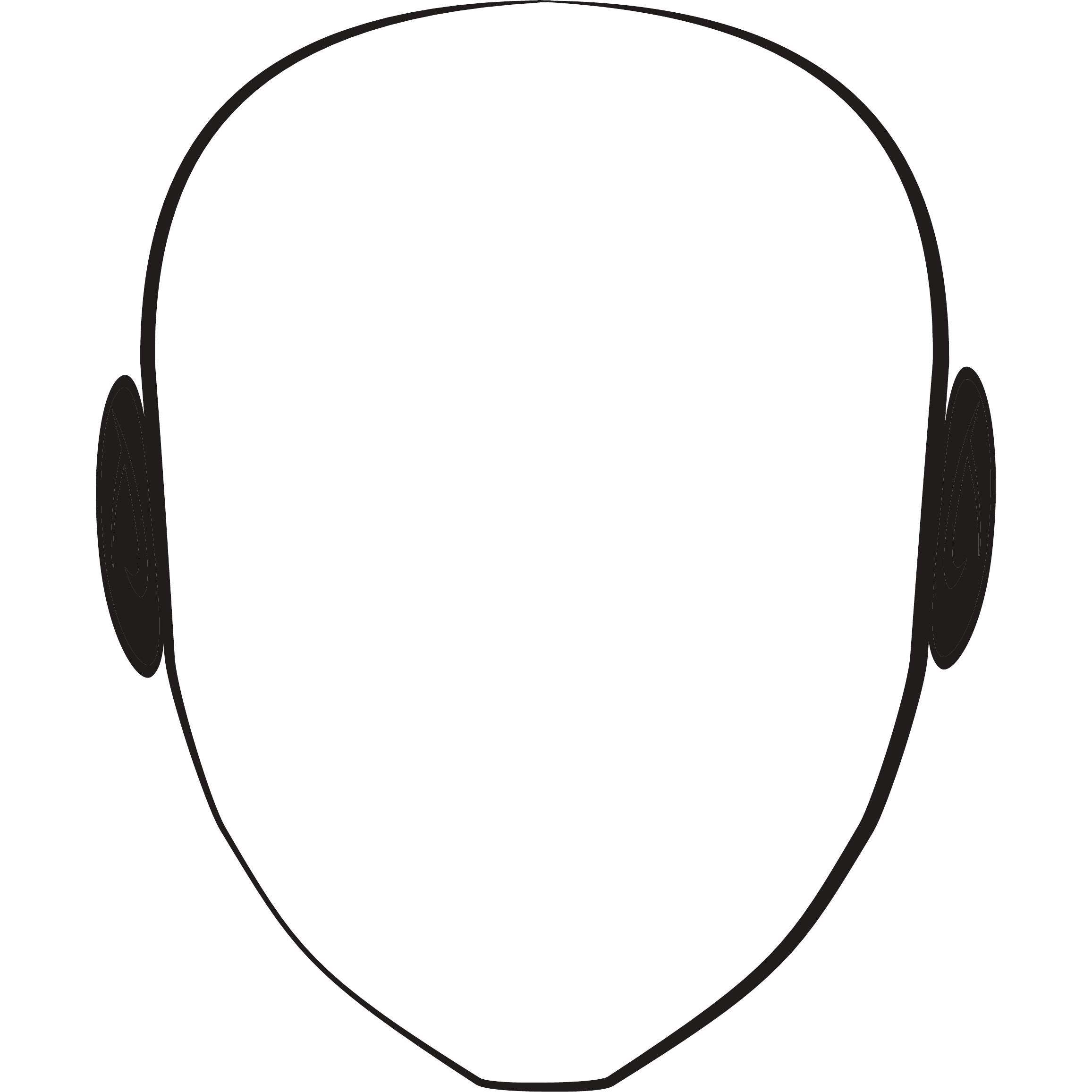} & \includegraphics[width = 0.5cm, height = 0.5cm]{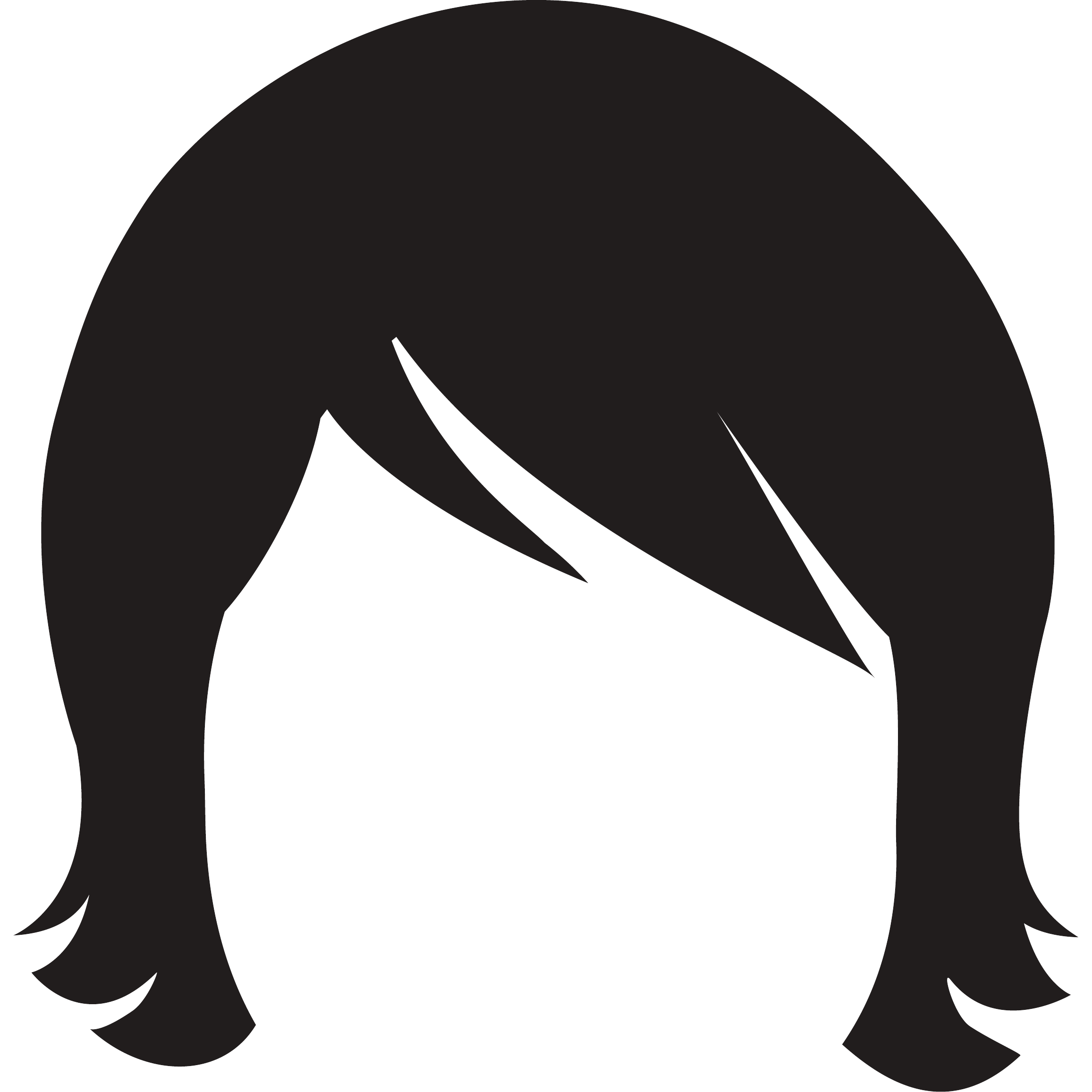} & \includegraphics[width = 0.5cm, height = 0.5cm]{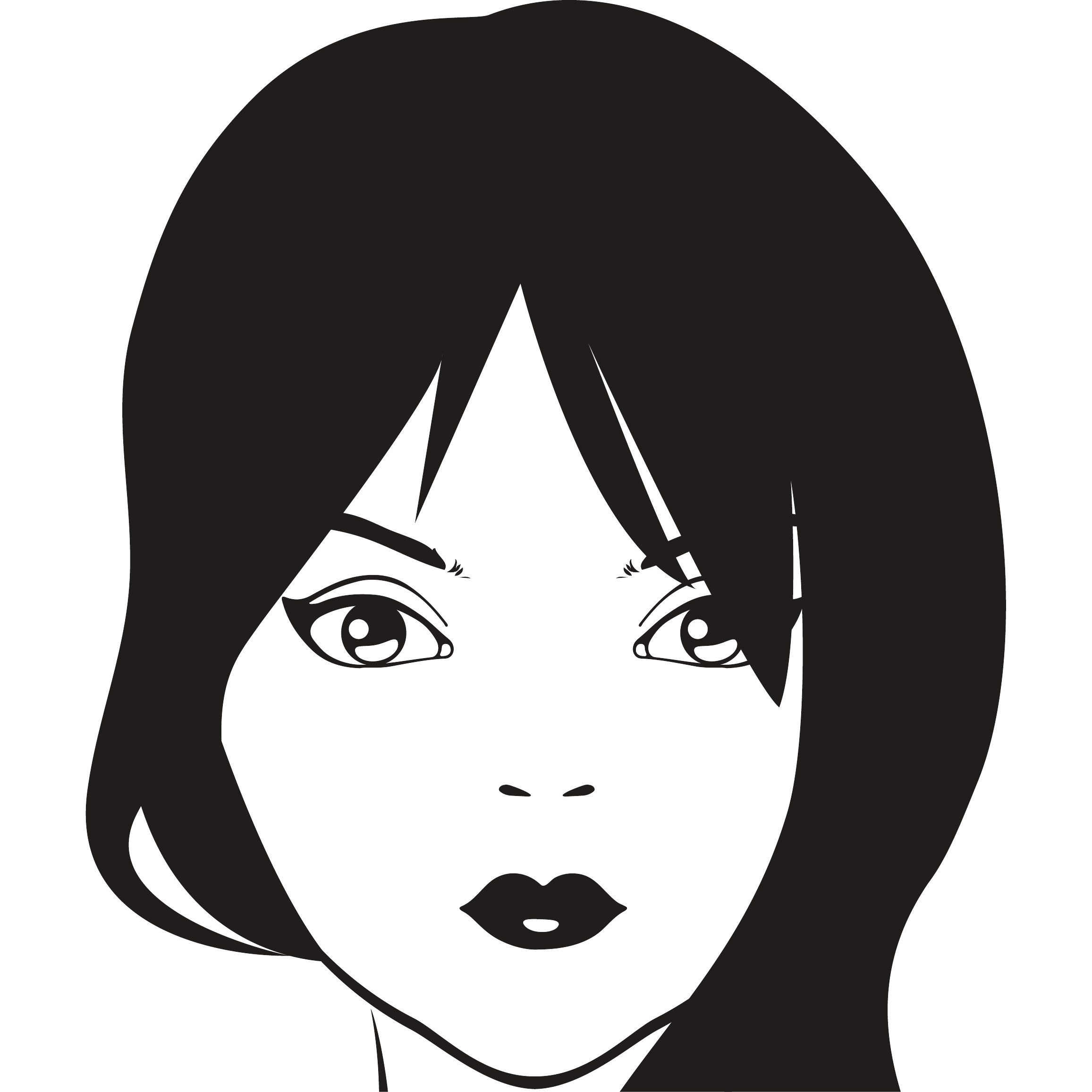} \\ \midrule
 & \includegraphics[width = 0.5cm, height = 0.5cm]{icons/background} & 97.97 & 00.73 & 01.30 & ----- \\
\multicolumn{1}{r}{\rot{\begin{tabular}[c]{@{}l@{}}true\\ class\end{tabular}}} & \includegraphics[width = 0.5cm, height = 0.5cm]{icons/face_skin} & 01.83 & 96.37 & 01.79 & ----- \\
 & \includegraphics[width = 0.5cm, height = 0.5cm]{icons/hair} & 06.35 & 05.44 & 88.21 & ----- \\ \midrule
\multicolumn{2}{c}{Jaccard index} & .9656 & .9182 & .7808 & .8882 \\ \bottomrule
\end{tabular}
\end{table}

\subsubsection{Helen dataset}

We iteratively trained the following five CnnRnnGan models for segmenting different classes:
\begin{itemize}
\item One global model for segmenting background, face skin and hair.
\item Three local models for segmenting eyebrows, eyes and nose, respectively.
\item One local model for segmenting upper lip, inner mouth and lower lip.
\end{itemize}

The outputs of the global model and the local models were aggregated by resizing the output of the global model to 500 pixels $\times$ 500 pixels and placing the non-background outputs of the local models to their corresponding locations in the resized output of the global model.

The global model was trained on the Helen dataset in the exact same way as it was trained on the Part Labels dataset. The local models were trained in a slightly different way than that in which the global models were trained. Before the first iteration, the images in the dataset were cropped to 90 pixels $\times$ 90 pixels such that their centers coincided with the centers of the corresponding classes of the average face. At each iteration, a mini-batch of size 16 was randomly selected without replacement, rotated by $\pm$ 7.5 degrees, scaled by a factor of 1 $\pm$ 0.05, horizontally and vertically translated by $\pm$ 5 pixels, and randomly flipped in the left-right direction. Additionally, the initial segmentations were further randomly rotated by $\pm$ 0.75 degrees, scaled by a factor of 1 $\pm$ 0.005, and horizontally and vertically translated by $\pm$ 0.5 pixels. The additional data augmentation was used to further avoid overfitting the training set since the training set had a small overlap with the training set of the landmark detection model. Finally, the mini-batch was cropped to the central 80 pixels $\times$ 80 pixels.

In the test phase, the inputs were oversampled (i.e., center and corners) and mirrored (i.e., left-right direction). The outputs were placed to their corresponding locations in the original inputs and averaged. Table~\ref{table_3} shows the resulting confusion matrix and Jaccard index. The most common cause of errors was mislabeling the classes as face skin and background. The least common cause of errors was mislabeling the classes as eyes and nose. Importantly, when the non-background outputs of the local models were misclassified, they were almost always misclassified as the output of the global model and almost never as one another, which suggests that the simple post-hoc aggregation of the outputs of the global model and the local models was sufficient. All of the classes were segmented with a relatively high accuracy ($J = 0.7873$). Background and face skin were the most accurately segmented classes ($J_{b} = 0.9452$ and $J_{fs} = 0.8933$). Hair and upper lip were the least accurately segmented classes ($J_{h} = 0.6962$ and $J_{ul} = 0.6619$).

\begin{table}[]
\centering
\caption{\textbf{The results of the main experiment on the Helen dataset.} Confidence matrix is reported in terms of percentage. The rest of the results are reported in terms of Jaccard index (i.e. intersection over union) of the classes and their arithmetic mean, respectively.}
\label{table_3}
\begin{tabular}{@{}lccccccccccc@{}}
\toprule
 &  & \multicolumn{9}{c}{predicted class} & \multicolumn{1}{l}{} \\
 &  & \includegraphics[width = 0.5cm, height = 0.5cm]{icons/background} & \includegraphics[width = 0.5cm, height = 0.5cm]{icons/face_skin} & \includegraphics[width = 0.5cm, height = 0.5cm]{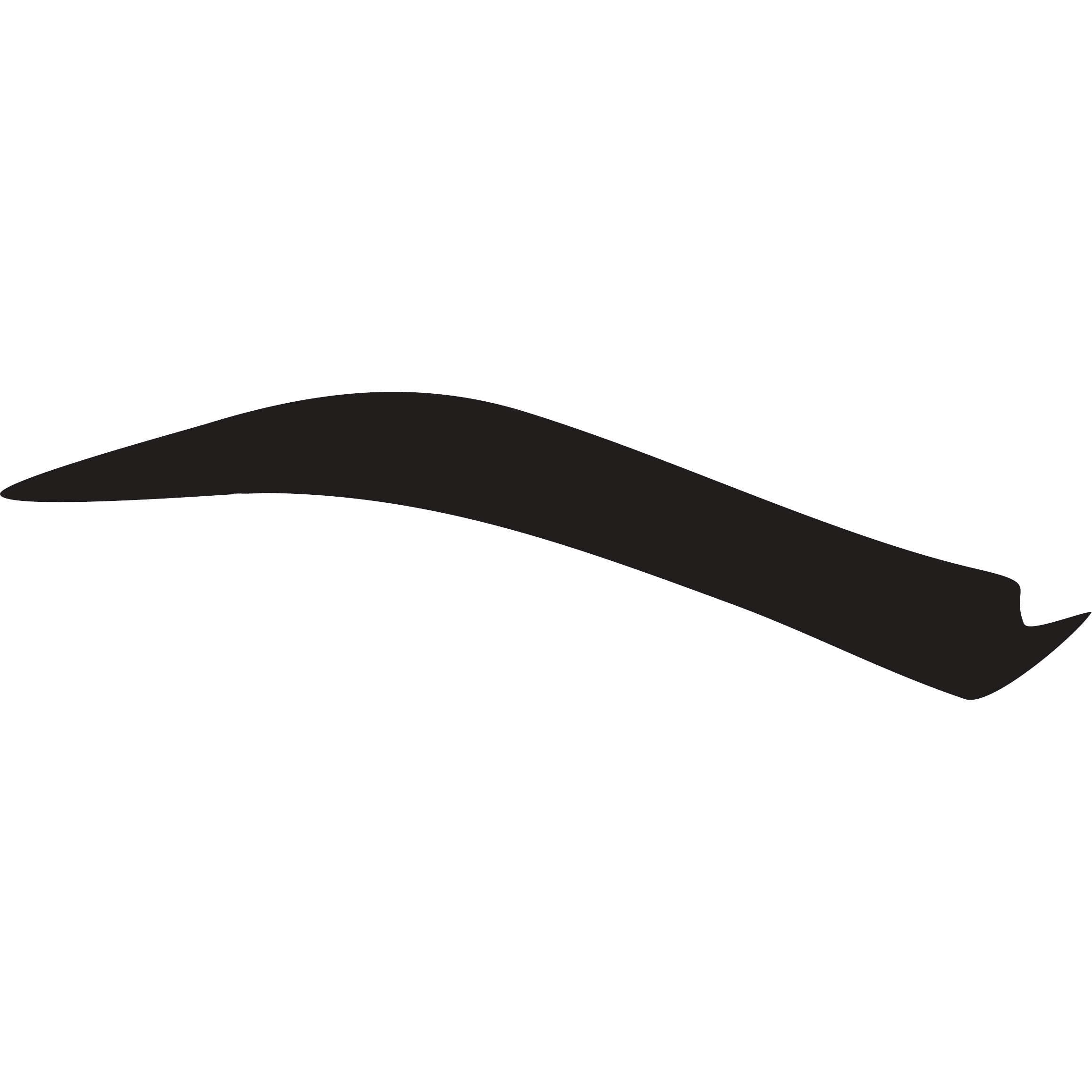} & \includegraphics[width = 0.5cm, height = 0.5cm]{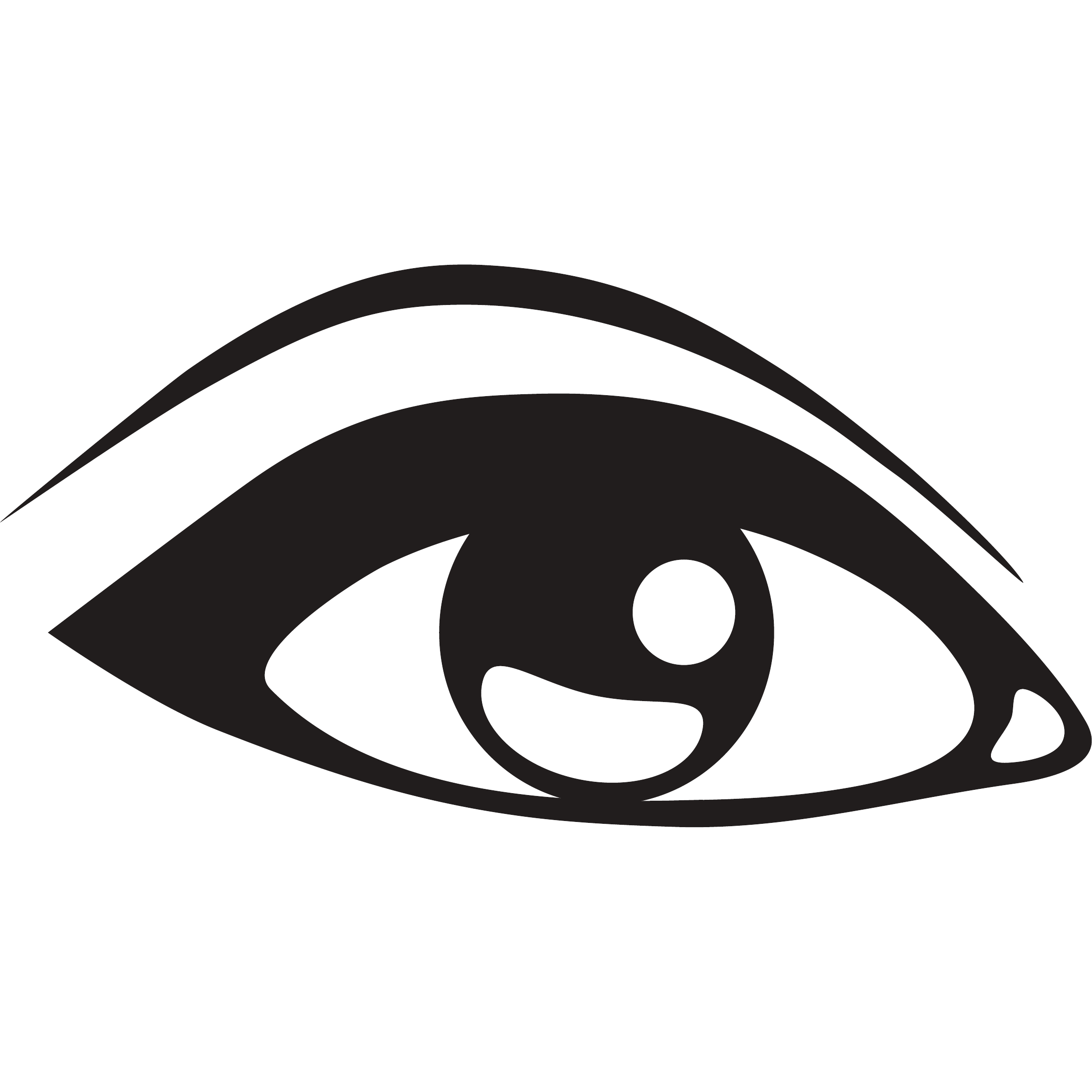} & \includegraphics[width = 0.5cm, height = 0.5cm]{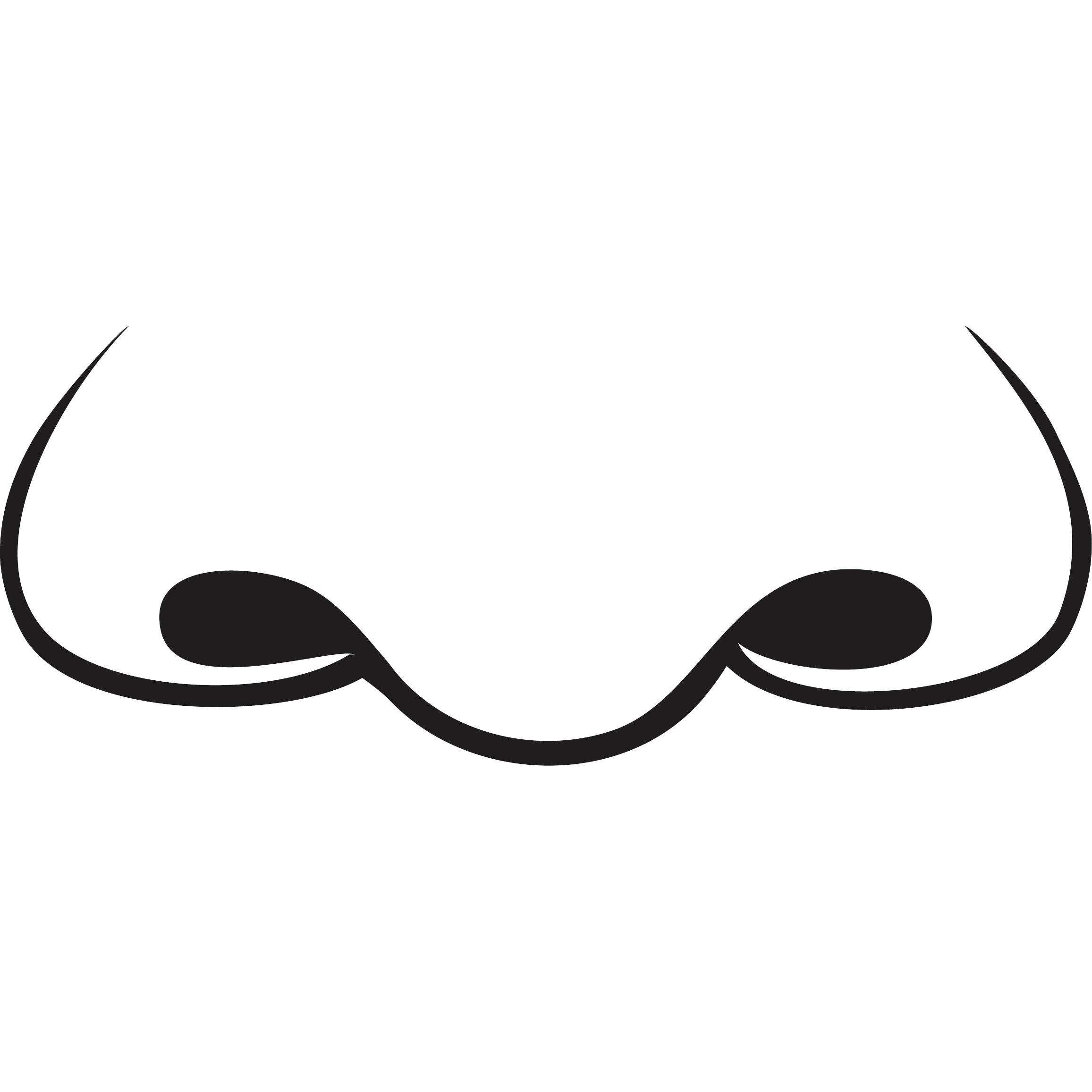} & \includegraphics[width = 0.5cm, height = 0.5cm]{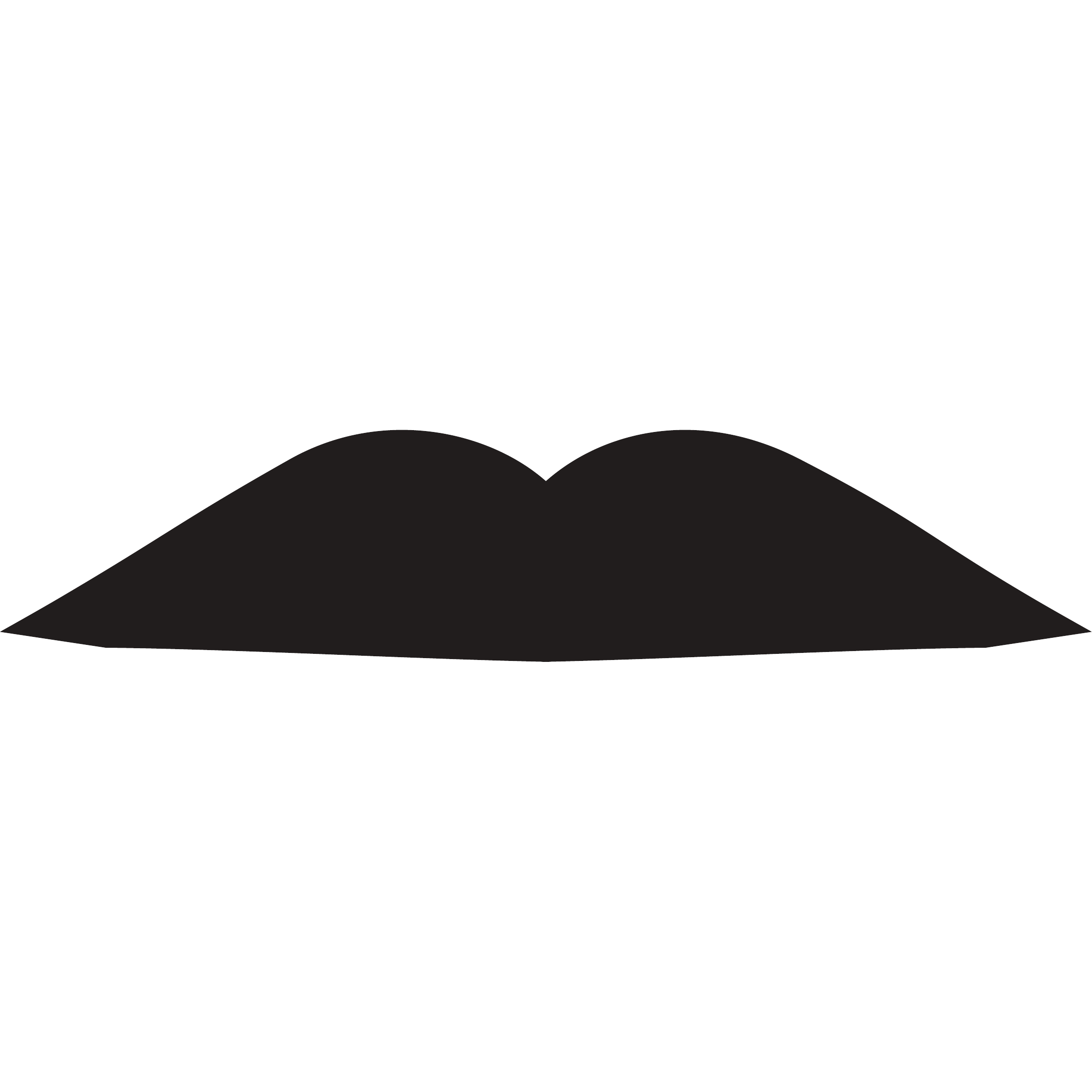} & \includegraphics[width = 0.5cm, height = 0.5cm]{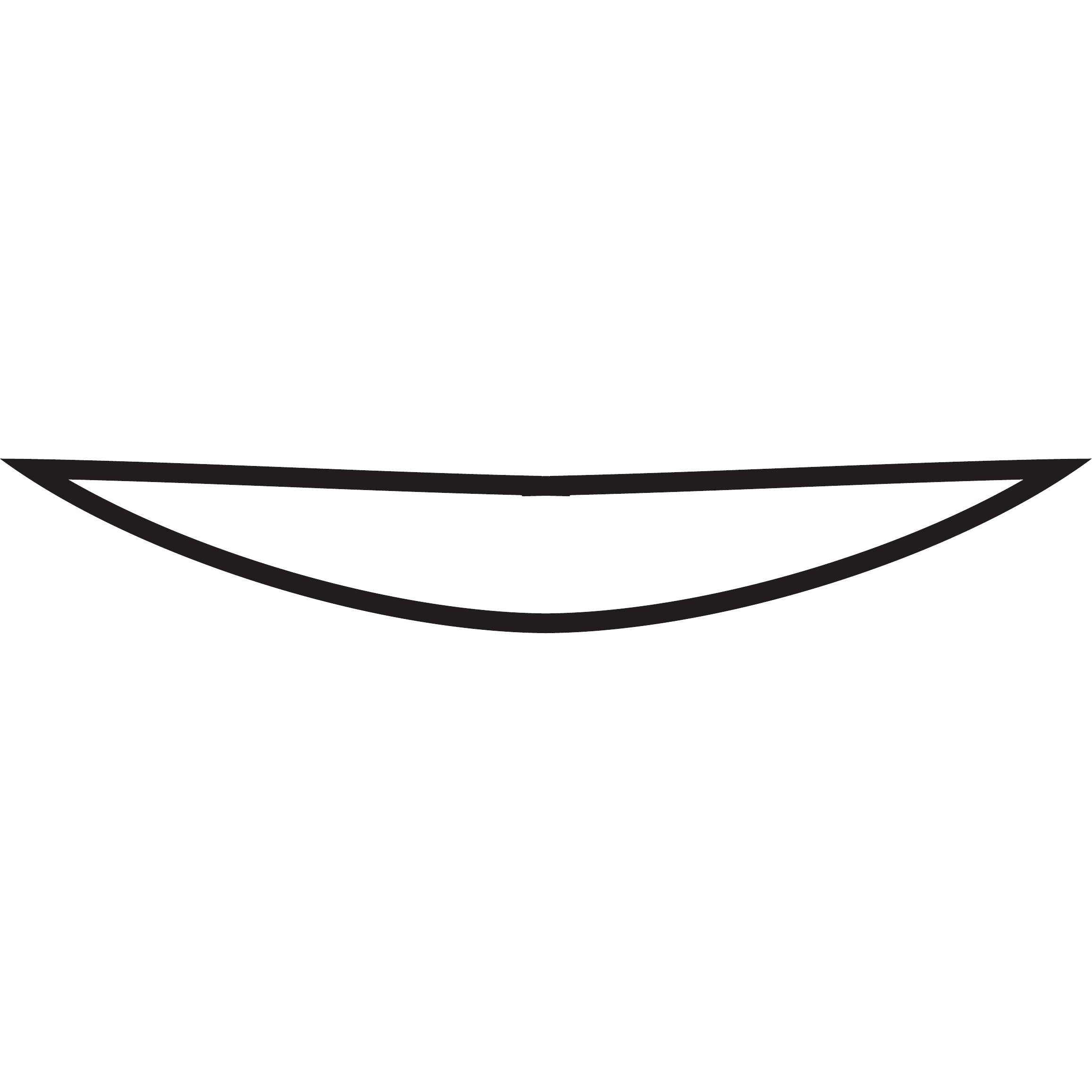} & \includegraphics[width = 0.5cm, height = 0.5cm]{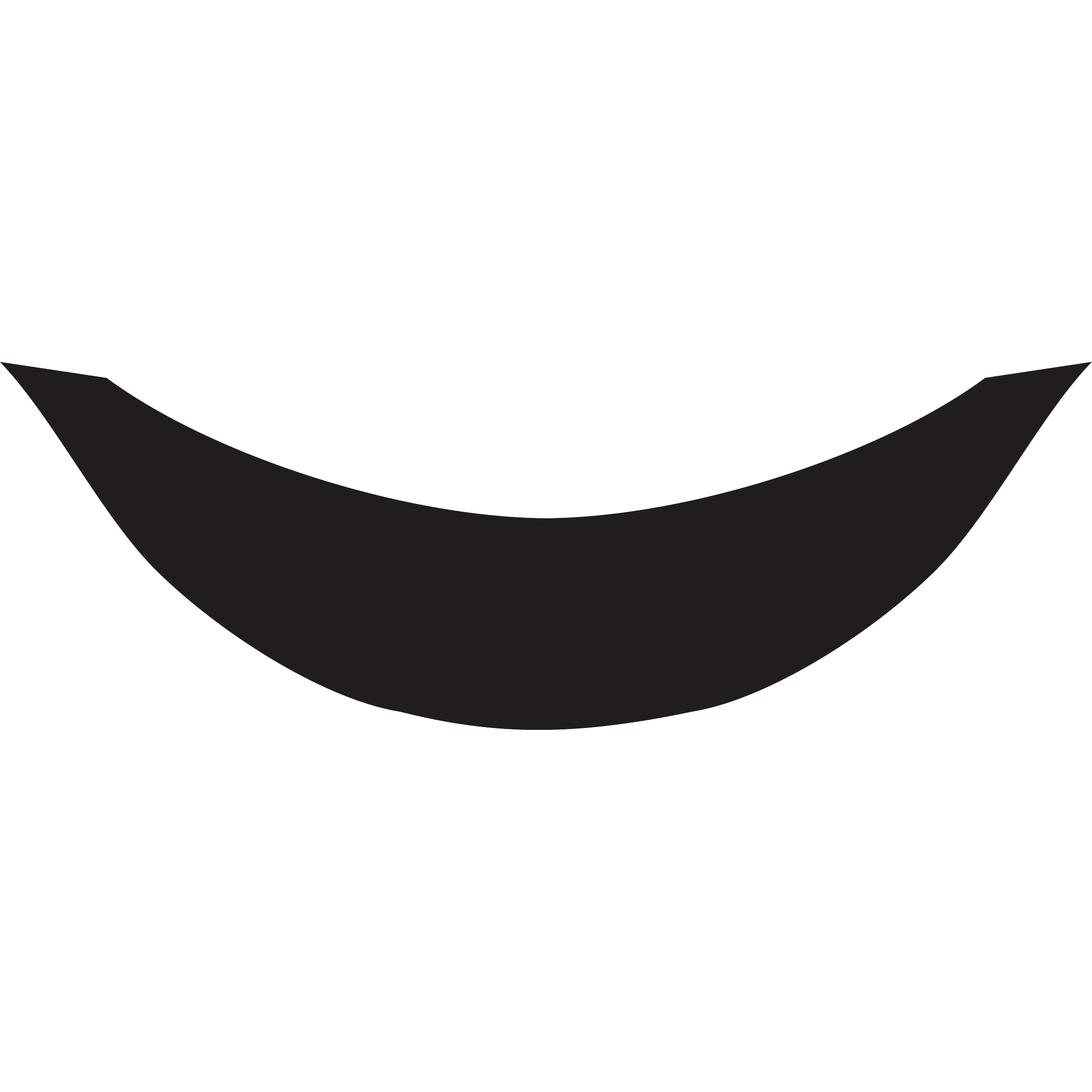} & \includegraphics[width = 0.5cm, height = 0.5cm]{icons/hair} & \includegraphics[width = 0.5cm, height = 0.5cm]{icons/overall} \\ \midrule
\multicolumn{1}{c}{} & \includegraphics[width = 0.5cm, height = 0.5cm]{icons/background} & 97.28 & 00.41 &  &  &  &  &  &  & 02.30 & ----- \\
\multicolumn{1}{r}{} & \includegraphics[width = 0.5cm, height = 0.5cm]{icons/face_skin} & 01.83 & 95.46 & 00.43 & 00.16 & 00.36 & 00.13 & 00.01 & 00.18 & 01.44 & ----- \\
 & \includegraphics[width = 0.5cm, height = 0.5cm]{icons/eyebrow} & 00.05 & 19.66 & 80.22 &  &  &  &  &  & 00.06 & ----- \\
\multicolumn{1}{r}{\multirow{3}{*}{\rot{\begin{tabular}[c]{@{}l@{}}true class\end{tabular}}}} & \includegraphics[width = 0.5cm, height = 0.5cm]{icons/eye} &  & 13.25 & 00.02 & 86.73 &  &  &  &  &  & ----- \\
\multicolumn{1}{r}{} & \includegraphics[width = 0.5cm, height = 0.5cm]{icons/nose} &  & 07.23 &  &  & 92.77 &  &  &  &  & ----- \\
\multicolumn{1}{r}{} & \includegraphics[width = 0.5cm, height = 0.5cm]{icons/upper_lip} &  & 14.16 &  &  & 00.01 & 80.90 & 03.63 & 01.30 &  & ----- \\
 & \includegraphics[width = 0.5cm, height = 0.5cm]{icons/inner_mouth} &  & 02.35 &  &  &  & 09.49 & 82.20 & 05.96 &  & ----- \\
 & \includegraphics[width = 0.5cm, height = 0.5cm]{icons/lower_lip} & 00.06 & 09.65 &  &  &  &  & 04.46 & 84.83 &  & ----- \\
 & \includegraphics[width = 0.5cm, height = 0.5cm]{icons/hair} & 16.38 & 02.70 & 00.11 &  &  &  &  &  & 80.81 & ----- \\ \midrule
\multicolumn{2}{c}{Jaccard index} & .9452 & .8933 & .6987 & .7974 & .8884 & .6619 & .7467 & .7580 & .6962 & .7873 \\ \bottomrule
\end{tabular}
\end{table}

Compared to the accuracy of hair segmentations on the Part Labels dataset, accuracy of hair segmentations on the Helen dataset was considerably lower ($J_h = 0.7808$ versus $J_h = 0.6962$). This discrepancy can be attributed to the way in which hair was annotated in the datasets. In the Part Labels dataset, hair was annotated by automatically segmenting images to superpixels and manually labeling the superpixels. In the Helen dataset, hair was automatically annotated by alpha matting.

In the Helen dataset, we observed relatively lower accuracy for hair, eyebrows and upper lips compared to the rest of the classes. The relative low accuracy of hair and eyebrows can be attributed to the fact that these classes do not have well defined boundaries making it difficult to isolate them from background and/or face skin. Similarly, the relatively  low accuracy of upper lip can be attributed to the fact that this class has shared borders with four other classes (i.e., face skin, inner mouth and lower lip) and often misclassified as belonging to one of them. However, the discrepancy between upper lip, and inner mouth or lower lip is surprising since these classes have the similar properties with upper lip, but might be explained by class imbalance.

\subsection{Comparison of results versus state-of-the-art}

After the main experiments, we compared the results of the CnnRnnGan model on the Part Labels dataset and the Helen dataset versus the earlier results reported in the literature.

\subsubsection{Part Labels dataset}

First, we compared our results on the Part Labels dataset versus the following:
\begin{itemize}
\item RBM and CRF based image labeling method of Kae et al. (2013)~\cite{kae2013augmenting}.
\item CNN, RBM and CRF based semantic part segmentation method of Tsogkas et al. (2015)~\cite{tsogkas2015semantic}.
\item CNN and CRF based face labeling method of Liu et al. (2015)~\cite{liu2015multi-objective}.
\item Convolutional VAE based semantic segmentation method of Zheng et al. (2015)~\cite{zheng2015learning}.
\item Convolutional neural fabric based semantic segmentation method of Saxena et al. (2016)~\cite{saxena2016convolutional}.
\end{itemize}

To the best of our knowledge, the CnnRnnGan model achieved state-of-the-art results on the Part Labels dataset (Table~\ref{table_4}). The best overall results in the literature~\cite{liu2015multi-objective, tsogkas2015semantic} were improved by 1.55 and 0.19 percentage points (pp) from 95.12 to 96.67 and from 96.97 to 97.16 for pixels and superpixels, respectively.\footnote{Note that the CnnRnnGan model was trained on pixels only. The superpixel results were obtained by averaging the corresponding outputs of the CnnRnnGan model. While these results are supoptimal since the CnnRnnGan model was not trained on superpixels, they are reported for completeness.} The improvements in the best existing hair results were more pronounced compared to those in the rest of the best existing results ($6.99$ pp versus $\leq 1.81$ pp).\footnote{Note that background, face skin and hair results were reported in~\cite{liu2015multi-objective} only.}

\begin{table}[]
\centering
\caption{\textbf{Comparison of our results versus the previous state-of-the-art on the Part Labels dataset.} The overall results are reported in terms of pixel and superpixel accuracy, respectively. The rest of the results are reported in terms of $F_1$ score.}
\label{table_4}
\begin{tabular}{@{}lccccc@{}}
\toprule
 & \includegraphics[width = 0.5cm, height = 0.5cm]{icons/background} & \includegraphics[width = 0.5cm, height = 0.5cm]{icons/face_skin} & \includegraphics[width = 0.5cm, height = 0.5cm]{icons/hair} & \multicolumn{2}{c}{\includegraphics[width = 0.5cm, height = 0.5cm]{icons/overall}} \\ \midrule
Kae et al. (2013)~\cite{kae2013augmenting} & ----- & ----- & ----- & ----- & 94.95 \\
Tsogkas et al. (2015)~\cite{tsogkas2015semantic} & ----- & ----- & ----- & ----- & 96.97 \\
Liu et al. (2015)~\cite{liu2015multi-objective} & 97.10 & 93.93 & 80.70 & 95.12 & ----- \\
Zheng et al. (2015)~\cite{zheng2015learning} & ----- & ----- & ----- & ----- & 96.59 \\
Saxena et al. (2016)~\cite{saxena2016convolutional} & ----- & ----- & ----- & 94.82 & 95.63 \\ \midrule
\textbf{Ours} & \textbf{98.25} & \textbf{95.74} & \textbf{87.69} & \textbf{96.67} & \textbf{97.16} \\ \bottomrule
\end{tabular}
\end{table}

\subsubsection{Helen dataset}

Second, we compared our results on the Helen dataset versus the following:
\begin{itemize}
\item Exemplar based face parsing method of Smith et. al (2013)~\cite{smith2013exemplar-based}.
\item CNN and CRF based face labeling method of Liu et al. (2015)~\cite{liu2015multi-objective}.
\item CNN based face parsing method of Liu et al. (2015)~\cite{zhou2015interlinked}.
\end{itemize}

To the best of our knowledge, our model achieved state-of-the-art results on the Helen dataset (Table~\ref{table_5}). The best overall result in the literature~\cite{zhou2015interlinked} was improved by 3.69 pp, from 87.30 to 90.99. The improvements in the best existing face skin, upper lip and lower lip results were more pronounced compared to those in the rest of the best existing results ($\geq 3.16$ pp versus $\leq 1.90$ pp).

\begin{table}[]
\centering
\caption{\textbf{Comparison of our results versus the state-of-the-art on the Helen dataset.} All of the results are reported in terms of $F_1$ score. The additional mouth results include the upper lip results, the inner mouth results and the lower lip results. The overall results exclude the background results and the face skin results.}
\label{table_5}
\begin{tabular}{@{}lccccc@{}}
\toprule
 & \includegraphics[width = 0.5cm, height = 0.5cm]{icons/face_skin} & \includegraphics[width = 0.5cm, height = 0.5cm]{icons/eyebrow} & \includegraphics[width = 0.5cm, height = 0.5cm]{icons/eye} & \includegraphics[width = 0.5cm, height = 0.5cm]{icons/nose} & ... \\ \midrule
Smith et. al (2013)~\cite{smith2013exemplar-based}& 88.20 & 72.20 & 78.50 & 92.20 & ... \\
Liu et. al (2015)~\cite{liu2015multi-objective} & 91.20 & 73.40 & 76.80 & 91.20 & ... \\ 
Zhou et. al (2015)~\cite{zhou2015interlinked} & ----- & 81.30 & 87.40 & \textbf{95.00} & ... \\ \midrule
Ours & \textbf{94.36} & \textbf{82.26} & \textbf{88.73} & 94.09 & ... \\ \midrule
 & \multicolumn{1}{l}{} & \multicolumn{1}{l}{} & \multicolumn{1}{l}{} & \multicolumn{1}{l}{} & \multicolumn{1}{l}{} \\ \midrule
 & \includegraphics[width = 0.5cm, height = 0.5cm]{icons/upper_lip} & \includegraphics[width= 0.5cm, height = 0.5cm]{icons/inner_mouth} & \includegraphics[width = 0.5cm, height = 0.5cm]{icons/lower_lip} & \includegraphics[width = 0.5cm, height = 0.5cm]{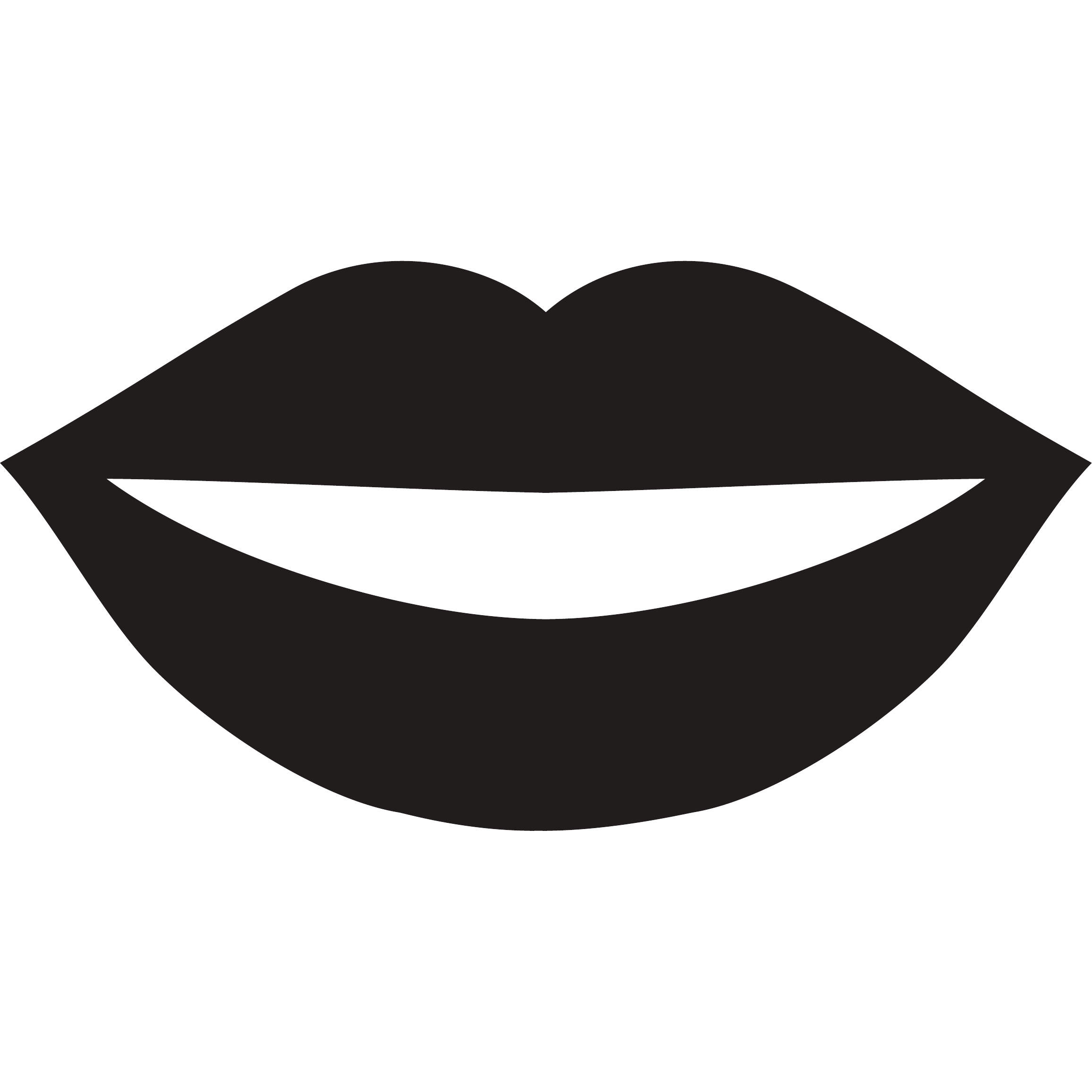} & \includegraphics[width = 0.5cm, height = 0.5cm]{icons/overall} \\ \midrule
Smith et. al (2013)~\cite{smith2013exemplar-based} & 65.10 & 71.30 & 70.00 & 85.70 & 80.40 \\
Liu et. al (2015)~\cite{liu2015multi-objective} & 60.10 & 82.40 & 68.40 & 84.90 & 85.40 \\
Zhou et. al (2015)~\cite{zhou2015interlinked} & 75.40 & 83.60 & 80.90 & 92.60 & 87.30 \\ \midrule
Ours & \textbf{79.66} & \textbf{85.50} & \textbf{86.23} & \textbf{92.82} & \textbf{90.99} \\ \bottomrule
\end{tabular}
\end{table}

\subsection{Ablation experiments}

Finally, we conducted two sets of ablation experiments on the Part Labels dataset and the Helen dataset, in which we evaluated the variants of the CnnRnnGan model.

\subsubsection{Part Labels dataset}

First, we evaluated the effect of removing the different components of the CnnRnnGan model on the Part Labels dataset (Table~\ref{table_6}).

The CnnRnnGan model achieved the best results except for background. The results were deteriorated by removing the Gan component and keeping the Rnn component (CnnRnn model). Removing the Rnn component and keeping the Gan component (CnnGan model) further decreased the accuracy. The results were once again deteriorated by removing both the Rnn component and the Gan component (Cnn model). Among all of the classes, the most notable change was observed for hair ($0.0283$).

\begin{table}[]
\centering
\caption{\textbf{The results of the ablation experiment on the Part Labels dataset.} The results are reported in terms of Jaccard index (i.e. intersection over union) of the classes and their arithmetic mean, respectively.}
\label{table_6}
\begin{tabular}{@{}lcccc@{}}
\toprule
 & \includegraphics[width = 0.5cm, height = 0.5cm]{icons/background} & \includegraphics[width = 0.5cm, height = 0.5cm]{icons/face_skin} & \includegraphics[width = 0.5cm, height = 0.5cm]{icons/hair} & \includegraphics[width = 0.5cm, height = 0.5cm]{icons/overall} \\ \midrule
Cnn & .9617 & .9111 & .7525 & .8751 \\
CnnGan & .9622 & .9114 & .7574 & .8770 \\
CnnRnn & \textbf{.9663} & .9177 & .7795 & .8878 \\ \midrule
\textbf{CnnRnnGan} & .9656 & \textbf{.9182} & \textbf{.7808} & \textbf{.8882} \\ \bottomrule
\end{tabular}
\end{table}

\begin{figure}
\centering
\includegraphics[width=1\textwidth]{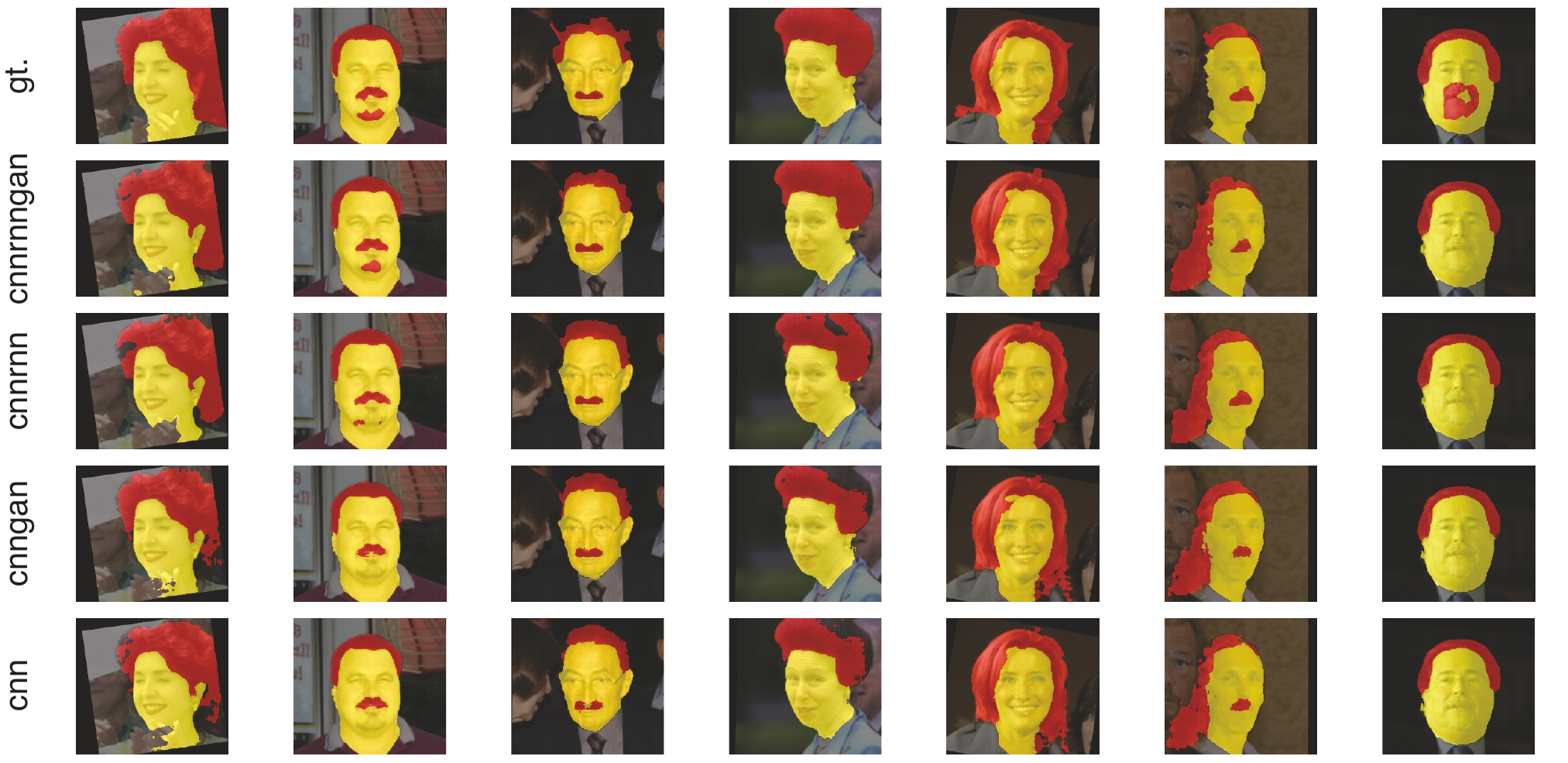}
\caption{\label{fig:qualitative_partlabels}\textbf{Example segmentations of the variants of the CnnRnnGan model that were evaluated in the ablation experiment on the Part Labels dataset.} The last two columns show failure cases in which none of the model variants achieved satisfactory results. gt. denotes ground-truth.}
\end{figure}

We illustrate qualitative examples of these results in Fig.~\ref{fig:qualitative_partlabels}. In the first column, it can be observed that even though the ground truth had a mistake (the hands were incorrectly labeled as face skin), particularly the CnnRnn and CnnRnnGan models correctly segmented most pixels. The example in the second column demonstrates the performance of the models in a difficult facial hair case. In this example, all models performed well in segmenting the mustache, but only CnnRnnGan model correctly identified the beard pixels. The third column showcases an example that all models performed well. The examples in the fourth and fifth columns highlight the gradual improvement provided by each additional model component in the correct classification of hair pixels. Especially in the example in the fifth column, it is possible to observe the improvements in the identification of fine details of hair. The first failure case example in column six demonstrates a difficult case for all models. The pixels to the left of the face skin are indeed hair pixels, however they belong to another person in the photograph. All models failed to make this distinction. The last failure case example shows that all models failed to segment the facial hair pixels and incorrectly labeled them as facial skin pixels. This error could be attributed to the low contrast difference between the face skin and facial hair pixels.

\subsubsection{Helen dataset}

Second, we evaluated the effect of conditioning the CnnRnnGan model on the initial segmentation and/or training multiple CnnRnnGan models for segmenting different classes on the Helen dataset (Table~\ref{table_7}). The initial segmentation (init. model) failed to achieve competitive results. These results were considerably improved by training a single CnnRnnGan model for segmenting all of the classes (1c. model). Conditioning the single CnnRnnGan model on the initial segmentation (init.+1c. model) slightly increased the accuracy. The results were once again considerably improved by training multiple CnnRnnGan models for segmenting different classes and conditioning them on the initial segmentation (init.+5c. model), which made them the best for all of the classes except for background and hair. Among all of the classes, the most notable improvements were observed for eyebrows, eyes, upper lip, inner mouth and lower lip ($\in [0.1051, 0.2132]$).

\begin{table}[]
\centering
\caption{\textbf{The results of the ablation experiment on the Helen dataset.} The results are reported in terms of Jaccard index (i.e. intersection over union) of the classes and their arithmetic mean, respectively.}
\label{table_7}
\begin{tabular}{@{}lcccccccccc@{}}
\toprule
 & \includegraphics[width = 0.5cm, height = 0.5cm]{icons/background} & \includegraphics[width = 0.5cm, height = 0.5cm]{icons/face_skin} & \includegraphics[width = 0.5cm, height = 0.5cm]{icons/eyebrow} & \includegraphics[width = 0.5cm, height = 0.5cm]{icons/eye} & \includegraphics[width = 0.5cm, height = 0.5cm]{icons/nose} & \includegraphics[width = 0.5cm, height = 0.5cm]{icons/upper_lip} & \includegraphics[width = 0.5cm, height = 0.5cm]{icons/inner_mouth} & \includegraphics[width = 0.5cm, height = 0.5cm]{icons/lower_lip} & \includegraphics[width = 0.5cm, height = 0.5cm]{icons/hair} & \includegraphics[width= 0.5cm, height = 0.5cm]{icons/overall} \\ \midrule
init. & .8253 & .6358 & .4855 & .6527 & .5325 & .5568 & .5757 & .6001 &  & .5405 \\
1c. & \textbf{.9465} & .8770 & .6074 & .6811 & .8562 & .5666 & .6655 & .6667 & \textbf{.7030} & .7300 \\
init.+1c. & .9408 & 8805 & .6189 & .6880 & .8618 & .5724 & .6804 & .6738 & .6717 & .7320 \\ \midrule
init.+5c. & .9452 & \textbf{.8933} & \textbf{.6987} & \textbf{.7974} & \textbf{.8884} & \textbf{.6619} & \textbf{.7467} & \textbf{.7580} & .6962 & \textbf{.7873} \\ \bottomrule
\end{tabular}
\end{table}

\begin{figure}
\centering
\includegraphics[width=1\textwidth]{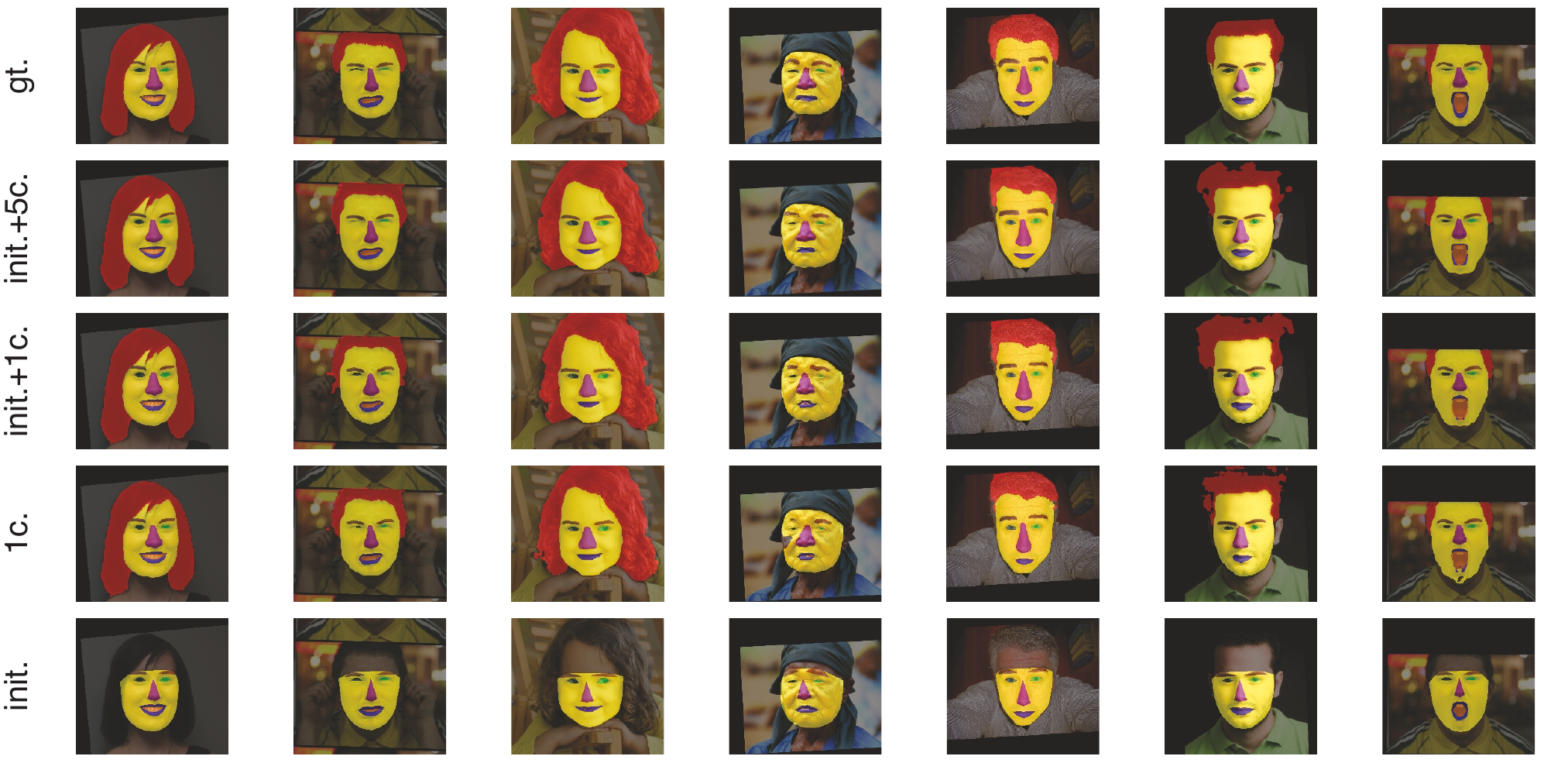}
\caption{\label{figure:figure_4}\textbf{Example segmentations of the variants of the CnnRnnGan model that were evaluated in the ablation experiment on the Helen dataset.} The last two columns show failure cases in which none of the model variants achieved satisfactory results. gt. denotes ground-truth.}
\end{figure}

We illustrate qualitative examples of these results in Fig.~\ref{figure:figure_4}. In the first five columns of this figure, it is possible to see an increase in performance starting from the simplest initial segmentation model to the complex variants of the CnnRnnGan model. While the initial segmentation does a good job in determining the general locations of each face region, it does not provide a detailed solution. Furthermore, it can be observed that the initial segmentation performs rather poorly in the nose and eyebrow regions, and whenever the expression of the face diverges from a neutral pose in the mouth regions. Among the variants of the CnnRnnGan model, the qualitative differences were minimal. However, the improvement provided by training multiple CnnRnnGan models for segmenting different classes and conditioning them on the initial segmentation (i.e. init.+5c) has resulted in visually distinguishable accuracy differences. This model was able to capture the details better than the remaining two model variants. The last two columns in the figure demonstrate failure cases where all model variants had errors. Models performed poorly in distinguishing hair from background when the background color was similar to the hair color (column 6) and in identifying the mouth regions when the person in the photograph had an extreme facial expression (column 7).

\section{Conclusion}
\label{section:conclusion}

Here, we proposed an end-to-end trainable semantic face segmentation model, which leverages the recent advances in the field. To this end, we formulated a conditional random field over a four-connected graph as convolutional and recurrent networks and estimated them via an adversarial process. Crucially, this formulation made it possible for this model to learn not only unary potentials but also pairwise potentials while aggregating multiscale contextual information and controlling higher-order inconsistencies. We showed that our model can exploit the structured nature of faces by conditioning it on face landmarks, and/or training it for different face landmarks and combining the outputs akin to part-based models. We evaluated our model on the Part Labels dataset and the Helen dataset, achieving state-of-the-art results on both of them while considerably improving the accuracy of challenging face parts such as hair. Future work will evaluate our model on other semantic segmentation datasets to asses its generalizability beyond faces.

\section*{Acknowledgements}
This work has been partially supported by VIDI grant number 639.072.513 of the Netherlands Organization for Scientific Research (NWO), the Spanish projects TIN2015-66951-C2-2-R, TIN2015-65464-R and TIN2016-74946-P (MINECO/FEDER, UE), by the European Comission Horizon 2020 granted project SEE.4C under call H2020-ICT-2015, and by the CERCA Programme/Generalitat de Catalunya.

\bibliographystyle{ieeetr}
\bibliography{main}

\end{document}